\documentclass{article}

\usepackage{microtype}
\usepackage{graphicx}
\usepackage{subfigure}
\usepackage{adjustbox}
\usepackage{booktabs} 
\usepackage{tabularx} 
\usepackage{ltablex}

\usepackage{hyperref}

\usepackage[accepted]{icml2024}

\usepackage{amsmath}
\usepackage{amssymb}
\usepackage{mathtools}
\usepackage{amsthm}
\usepackage{diagbox}
\usepackage{array}
\usepackage{multirow}
\usepackage{booktabs}      
\usepackage{array}         
\usepackage{makecell}      
\usepackage{graphicx}      
\usepackage{xcolor}   
\usepackage{colortbl} 
\usepackage{arydshln}
\usepackage[capitalize,noabbrev]{cleveref}
\usepackage{enumitem}

\theoremstyle{plain}

\theoremstyle{definition}

\theoremstyle{remark}

\usepackage[textsize=tiny]{todonotes}

\icmltitlerunning{Advancing Frontier Math Reasoning with Post-Training and Reward Modeling}

\begin{document}

\twocolumn[
\icmltitle{AceMath: Advancing Frontier Math Reasoning with Post-Training \\ and Reward Modeling}




\icmlsetsymbol{equal}{*}

\begin{icmlauthorlist}
\icmlauthor{Zihan Liu$^{\dagger}$}{equal,nvidia}
\icmlauthor{Yang Chen$^{\dagger}$}{equal,nvidia}
\icmlauthor{Mohammad Shoeybi}{nvidia}
\icmlauthor{Bryan Catanzaro}{nvidia}
\icmlauthor{Wei Ping$^{\dagger}$}{nvidia}
\end{icmlauthorlist}

\icmlaffiliation{nvidia}{NVIDIA}

\icmlcorrespondingauthor{Zihan Liu}{zihanl@nvidia.com}
\icmlcorrespondingauthor{Yang Chen}{yachen@nvidia.com}
\icmlcorrespondingauthor{\\Wei Ping}{wping@nvidia.com}

\icmlkeywords{Machine Learning, ICML}

\vskip 0.3in
]



\printAffiliationsAndNotice{\icmlEqualContribution} 

\begin{abstract}
In this paper, we introduce AceMath, a suite of frontier math models that excel in solving complex math problems, along with highly effective reward models capable of evaluating generated solutions and reliably identifying the correct ones.
To develop the instruction-tuned math models, we propose a supervised fine-tuning (SFT) process that first achieves competitive performance across general domains, followed by targeted fine-tuning for the math domain using a carefully curated set of prompts and synthetically generated responses.
The resulting model, \emph{AceMath-72B-Instruct} greatly outperforms Qwen2.5-Math-72B-Instruct, GPT-4o and Claude-3.5 Sonnet.
To develop math-specialized reward model, we first construct \emph{AceMath-RewardBench}, a comprehensive and robust benchmark for evaluating math reward models across diverse problems and difficulty levels.
After that, we present a systematic approach to build our math reward models. The resulting model, \emph{AceMath-72B-RM}, consistently outperforms state-of-the-art reward models. 
Furthermore, when combining AceMath-72B-Instruct with AceMath-72B-RM, we achieve the highest average rm@8 score across the math reasoning benchmarks.
We release model weights, training data, and evaluation benchmarks at: \url{https://research.nvidia.com/labs/adlr/acemath}.
\end{abstract}

\section{Introduction}
\label{sec:introduction}
\begin{figure*}[ht]
\includegraphics[width=\textwidth]{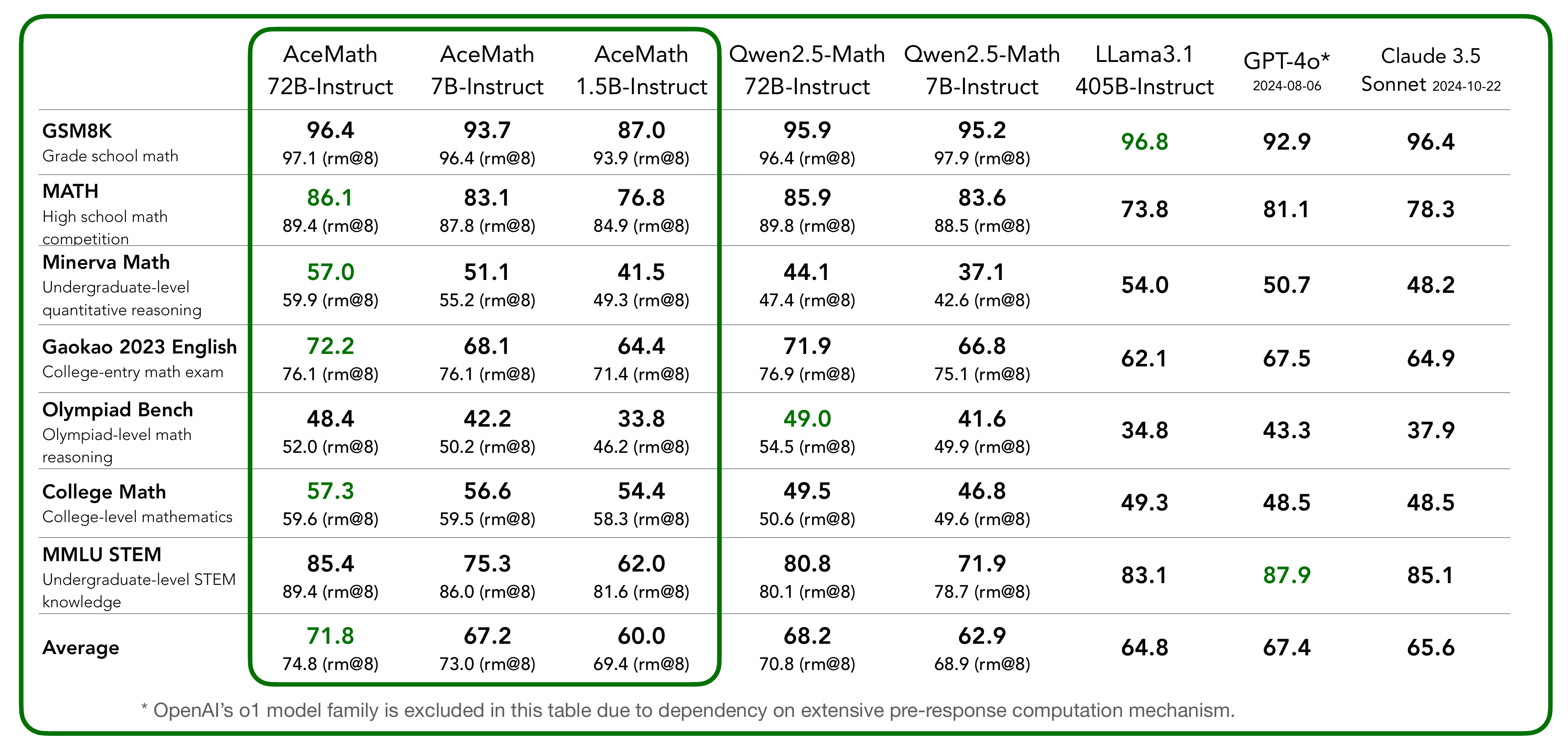}
\vspace{-0.6cm}
\caption{AceMath \textit{versus} leading open-weights and proprietary LLMs on math reasoning benchmarks. Additionally, we report rm@8 accuracy (best of 8) with our reward model AceMath-72B-RM and use the official reported numbers from Qwen2.5-Math.}
\label{fig:main_ressults}
\end{figure*}

Over the past year, the open large language model~(LLM) community has made remarkable progress in advancing the key capabilities of LLMs, including multi-turn conversation~\citep{vicuna2023, dubey2024llama}, coding~\citep{guo2024deepseekcoder, hui2024qwen2}, multimodal functionalities~\citep{dai2024nvlm, chen2024far}, retrieval-augmented generation (RAG)~\citep{liu2024chatqa}, and mathematical reasoning~\citep{azerbayev2023llemma, shao2024deepseekmath, Mathstral2024, yang2024qwen2_5_math}.
Among these capabilities, mathematics is recognized as a fundamental aspect of intelligence. It can serve as a reliable benchmark due to its objective, consistent, verifiable nature. Consequently, solving math problems is widely regarded as a critical testbed for evaluating an LLM's ability to tackle challenging tasks that require complex, numerical and multi-step logical reasoning~\citep[e.g.,][]{cobbe2021training, hendrycks2021measuring, lightman2023let}. 

Previous studies have convincingly demonstrated that math-specialized LLMs significantly outperform general-purpose LLMs on challenging mathematical benchmarks~\citep{azerbayev2023llemma, shao2024deepseekmath, Mathstral2024, yang2024qwen2_5_math}.
These math-specialized models, including the corresponding reward models~(a.k.a. verifiers), are not only valuable to the mathematics and science communities~\citep[e.g.,][]{embracing2024tao}, but they also provide valuable insights into data collection and serve as synthetic data generation tools, contributing to the advancement of future iterations of general-purpose LLMs.

The improved mathematical reasoning capabilities of math-specialized LLMs are generally acquired through both the continued pre-training and post-training:
\emph{i)} During continued pre-training stage, the models are initialized with \emph{general-purpose base} pretrained LLMs~(e.g., \emph{Llama-3.1-70B}~\citep{dubey2024llama}), and continually trained on extensive collections of mathematical corpora, often comprising hundreds of billions of tokens sourced from  Common Crawl~\citep{shao2024deepseekmath}, ArXiv papers~\citep{azerbayev2023llemma}, and synthetically generated datasets~\citep{yang2024qwen2_5_math, akter2024mind}. In this stage, losses are calculated on every token within the corpus.
\emph{ii)}  In the post-training phase, the continually pretrained \emph{math base} LLMs~(e.g., \emph{Qwen2.5-Math-72B}~\citep{yang2024qwen2_5_math}) are fine-tuned using large datasets of mathematical prompt-response pairs. In this stage, losses are computed only on the response tokens, allowing the models to refine their ability to generate accurate answers given the prompts or problem descriptions.

In this work, we push the limits of math reasoning with post-training and reward modeling based on open weights \emph{base} LLMs and \emph{math base} LLMs. 
We establish state-of-the-art supervised fine-tuning~(SFT) and reward modeling~(RM) processes for building math-specialized models, while also sharing key insights gained from our comprehensive studies.

Specifically, we make the following contributions: 
\begin{enumerate}[leftmargin=1.2em]
    \item We introduce a SFT process designed to first achieve competitive performance across general domains, including multidisciplinary topics, coding, and math.
    Building on this, the general SFT model is further fine-tuned in math domain using a meticulously curated set of prompts and synthetically generated responses.
    Leveraging the high-quality training data, the resulting model, AceMath-7B-Instruct, largely outperforms the previous best-in-class Qwen2.5-Math-7B-Instruct (pass@1: 67.2 vs. 62.9) on a variety of math reasoning benchmarks~(detailed results in Figure~\ref{fig:main_ressults}), while coming close to the performance of 10$\times$ larger Qwen2.5-Math-72B-Instruct (67.2 vs. 68.2). 
    Notably, our AceMath-72B-Instruct outperforms the state-of-the-art Qwen2.5-Math-72B-Instruct by a margin~(71.8 vs. 68.2).
    %
    \item We conducted a systematic investigation of training techniques for building math-specialized reward models, focusing on key aspects such as the construction of positive-negative pairs, training objectives, and the elimination of stylistic biases from specific LLMs.
    Leveraging the insights gained from this exploration, our AceMath-72B-RM consistently outperforms state-of-the-art reward models, including Qwen2.5-Math-RM-72B and Skywork-o1-Open-PRM-Qwen-2.5-7B~\citep{skyworkopeno12024}, in the math domain. 
    Moreover, when combining AceMath-72B-Instruct with AceMath-72B-RM, we achieve the highest average rm@8 score across seven math reasoning benchmarks~(see Figure~\ref{fig:main_ressults}), setting a new standard for performance in this field.
    \item We will open source the model weights for AceMath-Instruct and AceMath-RM, along with the complete training data used across all stages of their development.
    We also release AceMath-RewardBench, a comprehensive benchmark for evaluating math reward models, offering diverse datasets, varying difficulty levels, and robustness to variations in response styles.
\end{enumerate}

We organize the rest of this paper as follows. In \S\ref{sec:related_work}, we introduce the related work. 
In \S\ref{sec:sft}, we introduce the details on the SFT training data curation.
We present our reward model training in \S\ref{sec:reward_modeling}.
We conclude the paper in \S\ref{sec:conclusion}.

\section{Related Work}
\label{sec:related_work}

\subsection{Continued Pre-training on Math Corpus}

Many studies have investigated the integration of large-scale mathematical data for pre-training LLMs to enhance their math capabilities~\citep{shen2021mathbert,wang2023generative,zhang2024math,ying2024internlm,akter2024mind,hui2024qwen2}. 
Additionally, some research has focused on developing math-specialized LLMs by continuing the pre-training of a general-purpose LLM with an extensive math corpus, sourced from math-related web texts, encyclopedias, exam questions, and synthetic mathematical data~\citep{shao2024deepseekmath, yang2024qwen2_5_math}. These works demonstrate that this additional math-focused pre-training significantly enhances the model's ability to solve math problems, benefiting not only the pre-trained \emph{base} model but also subsequent \emph{instruct} models after post-training.

\subsection{Supervised Fine-Tuning}

Numerous supervised fine-tuning (SFT) datasets have been developed to enhance pretrained LLMs with versatile capability, such as instruction following~\citep{vicuna2023,ShareGPT_Vicuna_unfiltered,SlimOrca,mukherjee2023orca,GPTeacher2023,peng2023instruction,yuan2024advancing}, coding~\citep{GlaiveCodeAssistant2023,wei2024magicoder,luo2023wizardcoder}, and mathematical problem-solving~\citep{yue2023mammoth,yue2024mammoth2,yu2023metamath,mitra2024orca,numina_math_datasets}.
Due to the high cost of human-annotated data, synthetic data generation has become an essential component of SFT data construction, including both prompt and response augmentation~\citep{yu2023metamath,xu2024wizardlm,luo2023wizardcoder,li2024mugglemath,toshniwal2024openmathinstruct}.

Taking this further, math-instructed models have been developed to advance LLM performance in the mathematics domain~\citep{shao2024deepseekmath, toshniwal2024openmathinstruct, yang2024qwen2_5_math} by utilizing math-specific pretrained models as backbones and vast amounts of synthetic post-training data tailored to mathematics.
For example, OpenMathInstruct~\citep{toshniwal2024openmathinstruct} shows that math-specialized SFT with extensive synthetic data on the Llama3.1 base model significantly outperforms the corresponding Llama3.1 instruct model on mathematical benchmarks. In addition, Qwen2.5-Math~\citep{yang2024qwen2_5_math} demonstrates that a 7B math-instruct model can achieve math reasoning capabilities comparable to GPT-4o.





\subsection{Reward Modeling}
Training reward models for mathematical verification often involves discriminative approaches, such as binary classification to distinguish correct solutions from incorrect ones~\citep{cobbe2021training}. Alternatively, preference-based methods are employed, leveraging techniques like the Bradley-Terry loss~\citep{bt-1952,ouyang2022training} or regression loss to rank solutions, as demonstrated in models like HelpSteer~\citep{wang2024helpsteer2,wang2024helpsteer2preference}.
In contrast, generative reward models, such as LLM-as-a-judge~\citep{zheng2023judging} prompt LLMs to act as verifiers using predefined rubrics and grading templates \citep{bai2022constitutional}, GenRM \citep{zhang2024generative} leverages Chain-of-Thought reasoning~\citep{wei2022chain}, and Critic-RM~\citep{yu2024self} uses critic before predicting a reward. 
Our work on outcome reward model mainly focuses on robustness against style biases~\citep{liu2024rm} by sampling diverse model responses for training.
Beyond outcome-based reward models, process reward models (PRMs) provide step-by-step evaluations of model responses \citep{uesato2022solving, lightman2023let}. For example, Math-Shepherd \citep{wang-etal-2024-math} introduces an automated sampling method to construct large-scale process supervision data for training, following by further developments in step-wise supervision labeling~\citep{dong2024rlhf}, including PAV \citep{setlur2024rewarding}, OmegaPRM \citep{luo2024improve}, ER-PRM \citep{erprm}, AutoPSV \citep{lu2024autocv} and ProcessBench~\citep{processbench}.





\section{Supervised Fine-tuning}
\label{sec:sft}

\subsection{Overview}



Providing a strong initialization point is crucial for the model to begin math-focused SFT effectively.
Previous works~\citep{shao2024deepseekmath,yang2024qwen2_5_math} have demonstrated that continual pre-training of LLMs with a large math corpus provides a more effective initialization for subsequent math post-training. 
Taking this further, we explore whether conducting general SFT on pre-trained LLM can serves as a even better initialization for the subsequent math-specific SFT. 
The idea is that performing SFT on general-purpose tasks helps the model develop strong capabilities for following instructions and reasoning (e.g., knowledge-related). This foundation, in turn, makes it easier for the model to acquire math problem-solving skills from math-focused SFT data.
The details of curating general SFT data can be found in \S\ref{sec:general_sft}.

The next-step is constructing math-specific SFT data. It is crucial to develop a diverse set of math prompts accompanied by unified, step-by-step, and accurate solutions. The details of curating math SFT data can be found in \S\ref{sec:math_sft}. 

Figure~\ref{fig:tokens_math_coding_others} depicts the summary of the SFT data. The details of how we leverage general and math SFT data for the training can be found in \S\ref{sec:train_strategy}.

\subsection{Data Curation}

\subsubsection{General SFT Data}
\label{sec:general_sft}

Our goal is to build a general SFT model that serve as a strong starting point for the subsequent math-specific SFT. This general SFT model should excel at following instructions and answer a wide range of questions, including those related to math and coding.

\paragraph{Prompt Construction}
To achieve this goal, we collect prompts from a diverse range of open-source datasets, categorized as follows:

\begin{itemize}
    \item General domain: ShareGPT~\citep{vicuna2023, ShareGPT_Vicuna_unfiltered}, SlimOrca~\citep{SlimOrca,mukherjee2023orca}, EvolInstruct~\citep{xu2024wizardlm}, GPTeacher~\citep{GPTeacher2023}, AlpacaGPT4~\citep{peng2023instruction}, and UltraInteract~\citep{yuan2024advancing};
    \item Coding domain: Magicoder~\citep{wei2024magicoder}, WizardCoder~\citep{luo2023wizardcoder}, GlaiveCodeAssistant~\citep{GlaiveCodeAssistant2023}, and CodeSFT~\citep{adler2024nemotron};
    \item Math domain: NuminaMath~\citep{numina_math_datasets}, OrcaMathWordProblems~\citep{mitra2024orca}, MathInstruct~\citep{yue2023mammoth}, and MetaMathQA~\cite{yu2023metamath}, as well as our synthetic data (details in \S\ref{sec:math_sft}).
\end{itemize}

Since different data sources could have prompt overlaps, we conduct data deduplication to eliminate duplicate prompts that are identical when converted to lowercase.
After deduplication, we retain the prompt set unfiltered to preserve the diversity of prompts.

\paragraph{Response Construction}
After collecting the prompts, our goal is to construct high-quality responses in a consistent format so that models can learn more effectively. 
Therefore, we avoid using the original open-source responses for these prompts, as they may lack quality and have inconsistent formats due to being sourced from different curators or generated by different models.
We use GPT-4o-mini (2024-0718) to generate responses for collected prompts in coding and general domains.
GPT-4o-mini is selected for its strong performance across different tasks and instructions, as well as its compact size, which makes it both time-efficient and cost-efficient for producing a large volume of generated responses.
We put the details of constructing responses for math SFT prompts in \S\ref{sec:math_sft}.

We generate a single response for each prompt using greedy decoding, ultimately accumulating around 1.2 million coding SFT samples~(0.67 billion tokens) and 0.7 million samples~(0.55 billion tokens) in the general domain.
And, we take around 1.2 million samples (0.95 billion tokens) from the math SFT data (described in \S\ref{sec:math_sft}) for the general SFT.

\subsubsection{Math SFT Data}
\label{sec:math_sft}

The goal is to construct a diverse set of math prompts accompanied by unified, step-by-step, and accurate solutions.

\paragraph{Initial Prompts}
We first take math prompts from general SFT data, drawing specifically from open-source datasets: NuminaMath~\citep{numina_math_datasets}, OrcaMathWordProblems~\citep{mitra2024orca}, MathInstruct~\citep{yue2023mammoth}, and MetaMathQA~\cite{yu2023metamath}. These prompts cover a wide range of math problems, spanning grade school, high school, college-level, and Olympiad-level math challenges. After that, we perform data deduplication to remove duplicate prompts as before. Finally, we collect over 1.3 million initial prompts.

\paragraph{Synthetic Prompt Generation}
Furthermore, we generate additional synthetic prompts to enrich the diversity of our math prompt collection. 
This process involves two key steps: 1) leveraging diverse seed prompts to inspire a powerful \textit{instruct} model to generate entirely new, potentially more challenging or uncommon prompts, and 2) ensuring that the generated prompts are solvable, as unsolvable prompts can lead to incorrect answers, which may degrade performance when used for training.
Therefore, we select NuminaMath as our seed prompt source due to its broad coverage of math questions across various difficulty levels. Then, we apply two strategies inspired by~\citet{xu2024wizardlm}: \textit{in-breadth evolution} for generating more rare prompts and \textit{in-depth evolution} for generating more challenging ones. For synthetic prompt generation, we utilize GPT-4o-mini (2024-0718). 

It is crucial to filter out low-quality synthetic prompts. 
In particular, we find that one type of \textit{in-depth evolution}, which involves adding constraints to existing prompts to generate new ones, can sometimes produce unsolvable or overly challenging questions. This, in turn, may result in incorrect answers being included in the training data, ultimately degrading model performance (see ablation studies in \S\ref{sec:ablation_synthetic_data}).
As a result, we exclude this type of prompt augmentation. Moreover, we filter out the synthetic prompts exceeding 300 words, as excessively lengthy math-related prompts are often problematic or unsolvable. Finally, we refine the synthetic math prompts to approximately one million by filtering out 500K, ensuring a more curated dataset for training. As a result, we have a total collection of over 2.3 million math prompts (1.3M initial prompts + 1M synthetic prompts).
We provide the details about the synthetic prompt generation in Appendix~\ref{appendix:synthetic_prompt_generation}.

\paragraph{Response Construction}
We utilize Qwen2.5-Math-72B-Instruct for generating responses to math prompts, given its state-of-the-art performance across various math benchmarks.
We add the instruction, ``\textit{Please reason step by step, and put your final answer within \textbackslash\textbackslash $boxed\{\}$.}'' to the prompt to ensures the responses are presented in a clear, step-by-step format with a consistent style.

We generate a single response for each of the over 2.3M prompts and ensure consistency in the response format by selecting only those responses (along with their prompts) that adhere to a uniform structure (e.g., starting the response with a summary of the question and having the final answer within \textbackslash\textbackslash $boxed{}$).
Additionally, responses exceeding 2,500 words are excluded, along with their prompts, as excessive response length often indicates a verbose or incorrect solution, or an unfinished response.
Furthermore, while Qwen2.5-Math-72B-Instruct demonstrates strong capabilities, it occasionally produces repetitive strings (e.g., repeating the same text until reaching the maximum output length). We detect and remove such patterns, along with their corresponding prompts.
Although these cases represent only a small fraction of the dataset, they can negatively impact the final performance and are carefully filtered during the curation process.
After filtering, we obtain a total of around 2.3 million math SFT samples (1.83 billion tokens), of which around 1.2 million are utilized in the general SFT.

Qwen2.5-Math-72B-Instruct can still generate incorrect solutions which may negatively impact model training.
To mitigate this, we focus on identifying samples with accurate final answers to create a higher-quality dataset for training.

Our approach involves cross-checking answers generated by different models and treating solutions with consistent outcomes as highly likely to be correct. Specifically, we leverage another strong model, GPT-4o-mini (2024-0718), to generate responses. Since GPT-4o-mini is comparatively weaker in mathematics than Qwen2.5-Math-72B-Instruct, we generate two responses per prompt and consider answers consistent across both responses as potentially correct. Finally, we compare these answers with those from Qwen2.5-Math-72B-Instruct, and select matched final answers as high-quality solutions for training, which results in a total size of 800K math SFT samples.

\subsubsection{Data Decontamination}
Data decontamination is essential in SFT to ensure unbiased evaluation and to prevent models from memorizing test samples. Following~\citet{yang2024qwen2_5_math}, we conduct data decontamination for math SFT prompts. The process begins with text normalization and the removal of irrelevant punctuation for each math prompt. Next, we filter out the prompt that has a 13-gram overlap with the test data and the longest common subsequence exceeding 60\% of its length.
For the rest of non-math SFT prompts, we simply filter out those with a 13-gram overlap with test samples.

\subsection{Training Strategy}
\label{sec:train_strategy}

\subsubsection{General SFT Strategy}
Among general tasks, solving complex coding and math problems stands out as particularly challenging, and many general instruct models often struggle with them. To address this and develop a more effective general SFT model, we introduce a two-stage training approach. 

In stage-1, the model is trained on a large dataset specifically curated for code and math SFT tasks, providing a strong foundation in these areas. Stage-2 expands the scope by incorporating a balanced mix of code, math, and other general SFT data, broadening the model's capabilities and enhance the overall performance.

We organize the constructed general SFT data (around three million samples) to support this two-stage training. For stage-1, the majority of the coding and math samples are selected, leading to a total of around 2 million SFT samples. Stage-2 training utilizes the remaining coding and math SFT samples, a subset of the stage-1 data, along with all other general SFT samples, resulting in a total of around 1.6 million samples. For math SFT samples used in stage-2 training, we select only the cross-checked high-quality data where the final answers provided by GPT-4o-mini and Qwen2.5-Math-72B-Instruct align, as detailed in \S\ref{sec:math_sft}. This strategy ensures that stage-2 training integrates additional, diverse, and high-quality coding and math SFT samples, thereby fostering a more robust model.

\begin{table*}[t]
\centering
\resizebox{0.88\linewidth}{!}{
\begin{tabular}{lccccccc}
\toprule
Models      & HumanEval & MBPP  & GSM8K & MATH & MMLU & MMLU Pro &  Avg.  \\ \midrule


DeepSeek-Coder-7B-Instruct-v1.5  &  64.10 & 64.60 & 72.60 & 34.10 & 49.50 & - & - \\
AceInstruct-7B-DeepSeekCoder (Ours) & 78.05 & 73.54 & 82.56 & 55.62 & 54.65 & 33.28 & 62.95 \\ \midrule

Llama3.1-8B-Instruct & 72.60 & 69.60 & 84.50 & 51.90 & 69.40 & 48.30 & 66.05 \\
AceInstruct-8B-Llama3.1 (Ours) & 81.10 & 74.71 & 90.45 & 64.42 & 68.31 & 43.27 & \textbf{70.38} \\ \midrule

Qwen2.5-1.5B-Instruct & 61.60 & 63.20 & 73.20 & 55.20 & 58.37 & 32.40 & 57.33 \\
AceInstruct-1.5B-Qwen2.5 (Ours) & 73.17 & 65.76 & 80.44 & 60.34 & 58.17 & 33.78 & \textbf{61.94} \\ \midrule

Qwen2.5-7B-Instruct & 84.80 & 79.20 & 91.60 & 75.50 & 74.51 & 56.30 & \textbf{76.99} \\
AceInstruct-7B-Qwen2.5 (Ours) & 85.37 & 74.32 & 93.10 & 76.40 & 74.68 & 54.50 & 76.40 \\ \midrule

Qwen2.5-72B-Instruct & 86.60 & 88.20 & 95.80 & 83.10 & 84.67 & 71.10 & \textbf{84.91} \\
AceInstruct-72B-Qwen2.5 (Ours) & 89.63 & 83.66 & 96.36 & 84.50 & 83.88 & 66.10 & 84.02 \\ \bottomrule

\end{tabular}
}
\caption{Results of our AceInstruct general SFT models. We apply our proposed two-stage training strategy to conduct SFT on various base models from DeepSeekCoder, Llama3.1, and Qwen2.5. These finetuned models are then compared against the corresponding instruct baselines that are built upon the same base models.}
\label{table:sft_general_full_results}
\end{table*}

\begin{table*}[t]
\centering
\resizebox{0.9\linewidth}{!}{
\begin{tabular}{lccccccc}
\toprule
Models      & HumanEval & MBPP  & GSM8K & MATH & MMLU & MMLU Pro &  Avg.  \\ \midrule

AceInstruct-8B-Llama3.1 & \textbf{81.10} & \textbf{74.71} & \textbf{90.45} & \textbf{64.42} & \textbf{68.31} & \textbf{43.27} & \textbf{70.38} \\
\quad $\triangleright$ Single-Stage SFT w/ all general SFT data  & 78.66 & 69.26 & 87.79 & 56.80 & 67.62 & 42.64 & 67.13 \\ 
\quad $\triangleright$ Single-Stage SFT w/ only stage-2 data  &  73.78 & 67.32 & 88.17 & 55.84 & 67.48 & 42.85 & 65.91 \\  \midrule

AceInstruct-7B-Qwen2.5 & \textbf{85.37} & 74.32 & \textbf{93.10} & \textbf{76.40} & \textbf{74.68} & \textbf{54.50} & \textbf{76.40} \\ 
\quad $\triangleright$ Single-Stage SFT w/ all general SFT data  & 83.54  & \textbf{75.49}  & 91.96  & 75.04  & 73.96  & 53.36  & 75.56  \\
\quad $\triangleright$ Single-Stage SFT w/ only stage-2 data & 83.54 & 73.15 & 92.27 & 75.12 & 74.26 & 53.06 & 75.23 \\ \bottomrule

\end{tabular}
}
\caption{Ablation studies of our general SFT models regarding the effectiveness of the two-stage training strategy.}
\label{table:sft_general_ablation}
\end{table*}

\begin{figure}[t]
    \centering
    \includegraphics[width=0.99\linewidth]{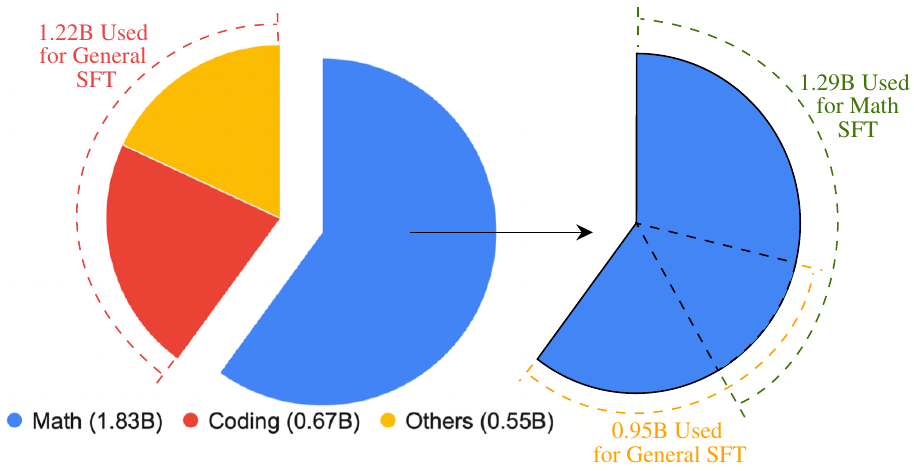}
    \caption{The proportion of total SFT tokens for math, coding, and other categories.}
    \label{fig:tokens_math_coding_others}
\end{figure}

\subsubsection{Math SFT Strategy}
\label{sec:train_strategy_math_sft}
We take the base (or math-base) model trained on our general SFT data as the starting point for the math SFT. In order to achieve diverse and high-quality math SFT data, we merge all samples from NuminaMath~\citep{numina_math_datasets}, a subset of samples from our synthetic prompts, and the 800K math SFT samples that are cross-checked between GPT-4o-mini and Qwen2.5-Math-72B-Instruct (as described in \S\ref{sec:math_sft}). We remove duplicate samples with identical prompts, resulting in a total of 1.6 million samples for math SFT. We find that this training blend leads to better results than directly utilize all 2.3 million math SFT samples for training (this ablation study can be found in \S\ref{sec:math_sft_training_data_ablation}).

\subsubsection{SFT Data Summary}
Figure~\ref{fig:tokens_math_coding_others} provides an overview of the distribution of total SFT tokens across math, coding, and other categories, along with details on the utilization of math SFT samples. 
In total, there are approximately 2.3 million math SFT samples (1.83 billion tokens), 1.2 million coding SFT samples (0.67 billion tokens), and 0.7 million samples in other categories (0.55 billion tokens). Among the math SFT samples, 1.2 million (0.95 billion tokens) are used for general SFT, while 1.6 million (1.29 billion tokens) are utilized for math SFT.

\subsubsection{Training Details}
All SFT models are trained using the AdamW optimizer~\citep{kingma2014adam,loshchilov2017decoupled}. We use a learning rate of 5e-6 for the general SFT and 3e-6 for the math SFT. A global batch size of 128 is used across all model sizes, except for the 72B model, where it is increased to 256. We conduct one epoch of training with a maximum sequence length of 4096 for both general SFT and math SFT.


\begin{table*}[t]
\centering
\resizebox{0.88\linewidth}{!}{
\begin{tabular}{lcccccccc}
\toprule
Models                                 & GSM8K & MATH  & \begin{tabular}[c]{@{}c@{}}Minerva\\ Math\end{tabular} & \begin{tabular}[c]{@{}c@{}}GaoKao\\ 2023 En\end{tabular} & \begin{tabular}[c]{@{}c@{}}Olympiad\\ Bench\end{tabular} & \begin{tabular}[c]{@{}c@{}}College\\ Math\end{tabular} & \begin{tabular}[c]{@{}c@{}}MMLU\\ STEM\end{tabular} & Avg.  \\ \midrule
GPT-4o (2024-0806)          & 92.90 & 81.10 & 50.74   & 67.50     & 43.30  & 48.50  & \textbf{87.99}    & 67.43 \\
Claude-3.5 Sonnet (2024-1022)      & 96.40 & 75.90 & 48.16 & 64.94 & 37.93 & 48.47 & 85.06 & 65.27 \\
Llama3.1-70B-Instruct & 94.10 & 65.70 & 34.20 & 54.00 & 27.70 & 42.50 & 80.40 & 56.94 \\
Llama3.1-405B-Instruct & \textbf{96.80} & 73.80 & 54.04 & 62.08 & 34.81 & 49.25 & 83.10 & 64.84 \\
OpenMath2-Llama3.1-8B   & 91.70 & 67.80 & 16.91 & 53.76 & 28.00 & 46.13 & 46.02 & 50.08 \\
Qwen2.5-Math-1.5B-Instruct    & 84.80 & 75.80 & 29.40   & 65.50  & 38.10   & 47.70    & 57.50   & 56.97 \\
Qwen2.5-Math-7B-Instruct      & 95.20 & 83.60 & 37.10  & 66.80 & 41.60  & 46.80   & 71.90  & 63.29 \\
Qwen2.5-Math-72B-Instruct     & 95.90 & 85.90 & 44.10   & 71.90   & \textbf{49.00}   & 49.50   & 80.80   & 68.16 \\  

\midrule
AceMath-1.5B-Instruct (Ours) & 86.95 & 76.84 & 41.54  & 64.42   & 33.78    & 54.36  & 62.04    & 59.99 \\ 
AceMath-7B-Instruct (Ours)   & 93.71 & 83.14 & 51.11  & 68.05   & 42.22   & 56.64  & 75.32  & 67.17 \\ 
AceMath-72B-Instruct (Ours)  & 96.44 & \textbf{86.10} & \textbf{56.99}   & \textbf{72.21}  & 48.44   & \textbf{57.24}   & 85.44    & \textbf{71.84} \\ \bottomrule
\end{tabular}
}
\caption{Greedy decoding (pass@1) results of math instruct models on math benchmarks. Our AceMath-1.5B/7B/72B-Instruct models are built upon the Qwen2.5-Math-1.5B/7B/72B-base models. AceMath-72B-Instruct greatly surpasses the previous state-of-the-art math-instruct model, Qwen2.5-Math-72B-Instruct.}
\label{table:sft_math_full_results}
\end{table*}

\subsection{Benchmarks}
\label{sec:benchmarks}

\subsubsection{General SFT Benchmarks}
We evaluate our general SFT models on a diverse set of widely used benchmarks.
These benchmarks consist of coding tasks, including HumanEval~\citep{chen2021evaluating} and MBPP~\citep{austin2021program}, mathematical reasoning, including GSM8K~\citep{cobbe2021training} and MATH~\citep{hendrycksmath2021}, as well as general knowledge domains, including MMLU~\citep{hendrycksmeasuring} and MMLU Pro~\citep{wang2024mmlu}. We conduct standard 5-shot evaluations for MMLU and MMLU Pro, and use 0-shot evaluations for the remaining benchmarks.


\subsubsection{Mathematical Benchmarks}
\label{sec:math_sft_benchmark}
We follow the evaluation setting in Qwen2.5-Math~\citep{yang2024qwen2_5_math} for assessing English mathematical tasks.
Beyond the commonly used GSM8K~\citep{cobbe2021training} and MATH~\citep{hendrycksmath2021} benchmarks, we also evaluate our models on a broader set of mathematical benchmarks, including Minerva Math~\citep{lewkowycz2022solving}, GaoKao 2023 En~\citep{liao2024mario}, Olympiad Bench~\citep{he2024olympiadbench}, College Math~\citep{tang2024mathscale}, and MMLU STEM~\citep{hendrycksmeasuring}. These benchmarks comprehensively assess a wide range of mathematical reasoning capabilities, from grade school arithmetic to advanced college-level problems and Olympic-level challenges.

Other than the above datasets, we further evaluate our models on AMC 2023\footnote{\url{https://huggingface.co/datasets/AI-MO/aimo-validation-amc}} and AIME 2024\footnote{\url{https://huggingface.co/datasets/AI-MO/aimo-validation-aime}}. Although these benchmarks are highly challenging math competition benchmarks, they are quite limited in size, with AMC 2023 containing only 40 test samples and AIME 2024 comprising just 30. Following~\citet{yang2024qwen2_5_math}, we evaluate these benchmarks separately and present the results in Appendix~\ref{appendix:amc_aime_results}.

We conduct 5-shot evaluations for MMLU STEM, and use 0-shot evaluations for the remaining benchmarks. 

Note that for all benchmarks except for Math and GSM8K, we do not use any training dataset or synthetic dataset derived from it. This ensures a more reliable and valid evaluation of our models on these benchmarks.


\subsection{Results of AceInstruct General SFT Models}

\subsubsection{Main Results}
As shown in Table~\ref{table:sft_general_full_results}, we apply our proposed two-stage training strategy to conduct SFT on various base models, including DeepSeekCoder-7B~\citep{guo2024deepseekcoder}, Llama3.1-8B~\citep{dubey2024llama}, and Qwen2.5-1.5B/7B/72B~\citep{yang2024qwen2}. We compare our finetuned general AceInstruct models to the corresponding instruct baselines that are built upon the same base models.
We observe that our general SFT brings significant improvements across different models, such as DeepSeek-Coder-7B, Llama3.1-8B, and Qwen2.5-1.5B, with an average score improvement of over 4\%. Notably, results on DeepSeek-Coder show that AceInstruct achieves particularly pronounced gains, with an average score increase of approximately 10\% or more in coding and math tasks. 
When compared to more advanced models like Qwen2.5-7B-Instruct and Qwen2.5-72B-instruct, our SFT delivers comparable performance. These findings highlight the effectiveness and strong generalization capabilities of our constructed general SFT dataset.

\subsubsection{Effectiveness of Two-Stage Training}
As shown in Table~\ref{table:sft_general_ablation}, we study the effectiveness of two-stage training strategy. For comparison, we use two base models from distinct families (Qwen2.5 and Llama3.1) and conduct single-stage training using either all general SFT data or only the stage-2 SFT data.

We observe that our two-stage training (AceInstruct) consistently outperforms single-stage training. 
Interestingly, we find notable improvements (more than 3\% average score) on a relatively weaker base model (e.g., Llama3.1-8B) compared to a stronger one (e.g., Qwen2.5-7B). This highlights the importance of incorporating extensive coding and math data during training to enhance the model's ability to handle complex coding and math tasks. 
We conjecture that the Qwen2.5 models already leverage substantial math and coding SFT data during pretraining, which reduces the effectiveness of an additional stage-1 SFT focused on these areas.

\begin{table*}[t]
\centering
\resizebox{0.84\linewidth}{!}{
\begin{tabular}{lcccccccc}
\toprule
Models                                 & GSM8K & MATH  & \begin{tabular}[c]{@{}c@{}}Minerva\\ Math\end{tabular} & \begin{tabular}[c]{@{}c@{}}GaoKao\\ 2023 En\end{tabular} & \begin{tabular}[c]{@{}c@{}}Olympiad\\ Bench\end{tabular} & \begin{tabular}[c]{@{}c@{}}College\\ Math\end{tabular} & \begin{tabular}[c]{@{}c@{}}MMLU\\ STEM\end{tabular} & Avg.  \\ \midrule

\multicolumn{9}{c}{\cellcolor[HTML]{CBF1F5}Backbone: Llama3.1-8B-Base} \\
AceMath-Instruct & 91.51 & 69.06 & 31.99  & 59.74  & 32.00  & 49.08  & 67.94  & 57.33 \\
\quad $\triangleright$  Only Qwen2.5-Math-72B-Instruct & 91.13 & 69.66 & 33.82 & 60.26 & 30.37 & 49.86 & 66.21 & 57.33 \\
\quad $\triangleright$  Only GPT-4o-mini & 90.83 & 68.12 & 36.03 & 60.26 & 31.70 & 48.05 & 66.50 & \textbf{57.36} \\ 
\quad $\triangleright$  Skipping general SFT & 91.81 & 68.70 & 31.99 & 59.48 & 31.11 & 48.40 & 62.76 & 56.32 \\ \midrule

\multicolumn{9}{c}{\cellcolor[HTML]{CBF1F5}Backbone: Qwen2.5-7B-Base} \\
AceMath-Instruct & 93.56 & 77.10 & 43.38 & 65.19 & 37.78 & 54.90 & 77.41 & \textbf{64.19} \\
\quad $\triangleright$  Only Qwen2.5-Math-72B-Instruct & 92.80 & 76.96 & 41.91 & 63.64 & 38.07 & 54.93 & 75.64 & 63.42 \\
\quad $\triangleright$  Only GPT-4o-mini & 91.66 & 74.14 & 43.75 & 64.42 & 39.26 & 52.27 & 76.03 & 63.08 \\
\quad $\triangleright$  Math SFT with all math samples  &  93.40  & 77.12 & 42.28 & 65.19 & 37.78 & 54.05 & 75.33 & 63.59  \\
\quad $\triangleright$  Math SFT with only cross-checked samples & 92.72 & 76.76 & 41.54 & 65.97 & 36.74 & 54.33 & 76.78 & 63.55 \\
\quad $\triangleright$  Skipping general SFT & 93.03 & 77.52 & 40.44 & 62.86 & 37.19 & 54.58 & 75.77 & 63.06 \\ \midrule


\multicolumn{9}{c}{\cellcolor[HTML]{CBF1F5}Backbone: Qwen2.5-Math-72B-Base} \\
AceMath-Instruct  & 96.44 & 86.10 & 56.99   & 72.21  & 48.44   & 57.24   & 85.44    & \textbf{71.84} \\
\quad $\triangleright$  Math SFT with all math samples & 96.29 & 86.06 & 55.15 & 70.13 & 46.67 & 57.49 & 84.96 & 70.96 \\
\quad $\triangleright$  Skipping general SFT & 95.75 & 85.52 & 56.25 & 71.43 & 45.33 & 56.71 & 84.42 & 70.77 \\ \bottomrule

\end{tabular}
}
\caption{Ablation Studies on training data and strategies across various backbone models for training our AceMath-Instruct models. The ablation studies can be categorized into three parts: 1) evaluating the effectiveness of using either GPT-4o-mini responses or Qwen2.5-Math-72B-Instruct responses individually; 2) analyzing the effectiveness of different math-specific samples for math SFT; and 3) assessing the impact of conducting general SFT prior to math SFT.}
\label{table:sft_math_ablation_training_data}
\end{table*}

\subsection{Results of AceMath-Instruct}

\subsubsection{Main Results}
In Table~\ref{table:sft_math_full_results}, we compare our AceMath-Instruct models against several strong baselines for greedy decoding, including Qwen2.5-Math-1.5B/7B/72B-Instruct~\citep{yang2024qwen2_5_math}, GPT-4o~\citep{gpt4o}, and Claude-3.5 Sonnet~\citep{claude3_5}. Specifically, our AceMath-1.5B/7B/72B-Instruct models are built upon the Qwen2.5-Math-1.5B/7B/72B-base models, which also serve as the foundation for Qwen2.5-Math-1.5B/7B/72B-Instruct. We find that AceMath-1.5B, 7B, and 72B-Instruct achieve significantly better performance compared to the corresponding Qwen2.5-Math-1.5B, 7B, and 72B-Instruct models. Our best model, AceMath-72B-Instruct, achieves a significant average improvement of 3.68 over the previous state-of-the-art, Qwen2.5-Math-72B-Instruct. This highlights the superior quality and generalizability of our constructed math SFT data. 

Moreover, we find that our 7B model, AceMath-7B-Instruct, demonstrate superior or comparable performance compared to several advanced instruct models, including Llama3.1-405B-Instruct, GPT-4o, and Claude-3.5 Sonnet. And, it comes close to matching the performance of the significantly larger Qwen2.5-Math-72B-Instruct, with only a slight difference in the average score (68.16 vs. 67.17).

We put several chain-of-thought reasoning examples generated by AceMath-72B-Instruct in Appendix~\ref{appendix:cot_examples}.

\begin{figure}[t]
    \centering
    \includegraphics[width=0.99\linewidth]{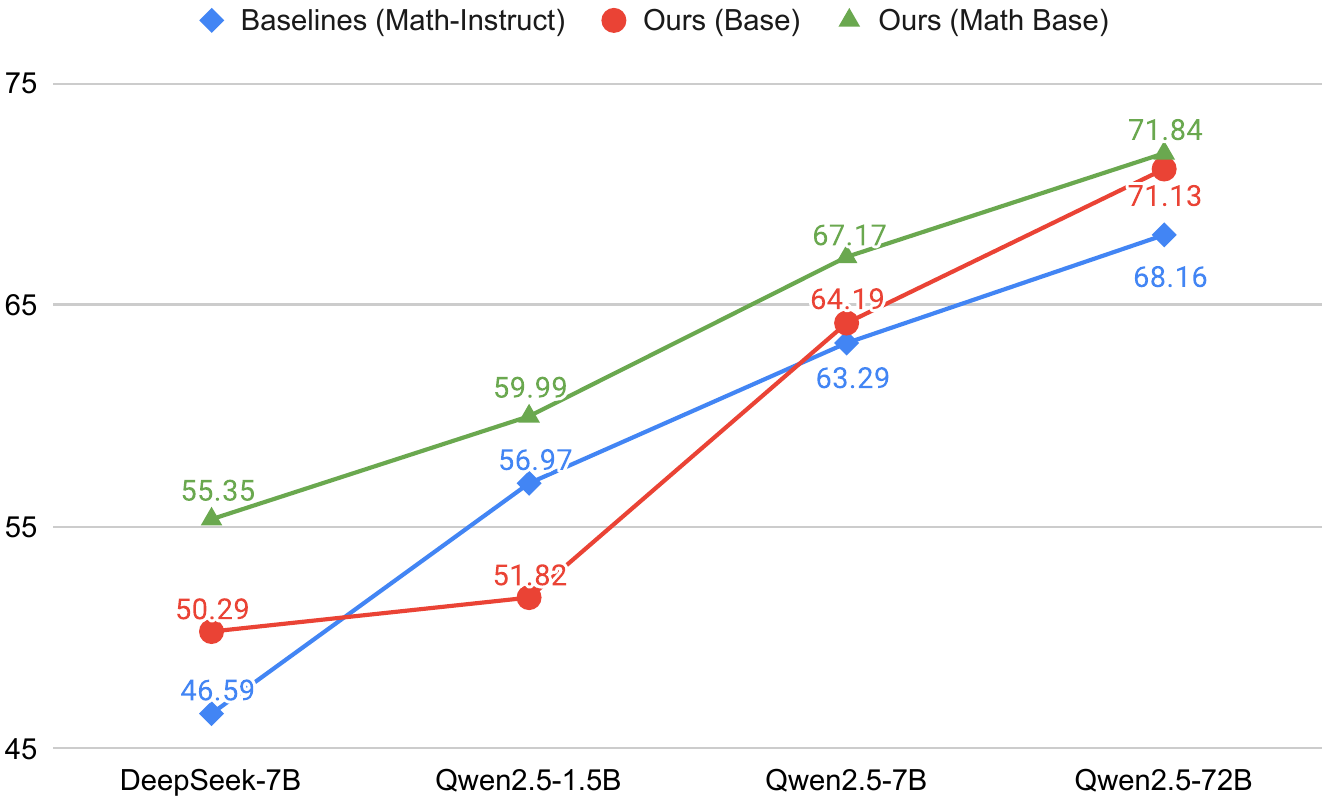}
    \caption{Studies on the impact of using either the base model or the math base model as the backbone on the performance of our AceMath-Instruct models. We compare our models against the corresponding math-instruct baselines across different model types and sizes. Results are the average scores of greedy decoding over the math benchmarks.}
    \label{fig:base_vs_mathbase}
\end{figure}

\subsubsection{Backbone Model: Base vs. Math-Base}
In Figure~\ref{fig:base_vs_mathbase}, we study the impact of using either the \textit{base} model (e.g., Qwen2.5-7B-Base) or the \textit{math base} model (e.g., Qwen2.5-Math-7B-Base) as the backbone on the performance of our AceMath-Instruct models. This study is crucial, as it helps us understand the importance of continual pre-training on a large math corpus (i.e., building \textit{math base} models) for improving the performance on solving math questions after post-training. 

For DeepSeek-7B, ``Ours (Base)'' uses the DeepSeek-Coder-7B-Base~\citep{guo2024deepseekcoder} as the backbone model, while ``Ours (Math Base)'' uses the DeepSeek-Math-7B-Base~\citep{shao2024deepseekmath} as the backbone model, which continues the pre-training of DeepSeek-Coder-7B-Base using a large math corpus. The math instruct baseline is DeepSeek-Math-7B-RL~\citep{shao2024deepseekmath}, which is developed from DeepSeek-Math-7B-Base. For Qwen2.5-1.5/7B/72B, the base models are Qwen2.5-1.5/7B/72B-Base, while the math base models are Qwen2.5-Math-1.5/7B/72B-Base, with the baselines being Qwen2.5-Math-1.5/7B/72B-Instruct.

We find that as the model size increases, the performance of our models with base models as backbones approaches that of models with math base as backbones. Specifically, when the Qwen2.5-(Math)-72B-Base is used, the performance gap between ``Ours (Base)'' and ``Ours (Math Base)'' becomes very marginal (71.84 vs. 71.13). We conjecture that larger models inherently possess better math problem-solving and generalization capability, which diminishes the need for continual pre-training.
This finding extends across different model families. Additionally, when comparing models of sizes between 1.5B and 7B, the performance gap between ``Ours (Base)'' and ``Ours (Math Base)'' is smaller for 7B models (i.e., DeepSeek-7B and Qwen2.5-7B) than it is for Qwen2.5-1.5B.

Moreover, we observe that except for Qwen2.5-1.5B, all the models from ``Ours (Base)'' outperform the corresponding math-instruct models that use stronger math base models as backbones. This further indicates that smaller models (e.g., 1.5B) rely more on continual pre-training with a large math corpus to enhance their math problem-solving capability (full results can be found in Appendix~\ref{appendix:full_base_mathbase_backbone}).

\begin{table}[t]
\centering
\resizebox{0.88\linewidth}{!}{
\begin{tabular}{lcccccccc}
\toprule
Models             & Average  \\ \midrule

AceMath-Instruct & \textbf{64.19} \\
\quad $\triangleright$ Removing all synthetic data  & 62.53 \\
\quad $\triangleright$ Using extra low-quality synthetic data  & 62.95 \\
\bottomrule

\end{tabular}
}
\caption{Ablation studies on the synthetic data, exploring the effects of removing all synthetic math SFT data and incorporating additional low-quality synthetic math SFT data. The backbone of AceMath-Instruct is Qwen2.5-7B-Base. Results are average across the seven math benchmark.}
\label{table:sft_math_ablation_synthetic}
\end{table}

\subsubsection{Ablation Studies on Training Strategy}
\label{sec:math_sft_training_data_ablation}
In Table~\ref{table:sft_math_ablation_training_data}, we conduct ablation studies on training data and strategies across various backbone models for training our AceMath-Instruct models.

First, we explore the effectiveness of using either GPT-4o-mini responses or Qwen2.5-Math-72B-Instruct responses individually. Given that our best-performing models leverage responses from both, we analyze the impact of relying solely on one model when constructing general and math SFT data. Notably, even when only GPT-4o-mini responses are available, we achieve strong performance, with just a 1\% average score drop when Qwen2.5-7B-Base serves as the backbone model.
Furthermore, with Llama3.1-8B-Base as the backbone, using responses from GPT-4o-mini, Qwen2.5-Math-72B-Instruct, or their combination (AceMath-Instruct) yields comparable results.
This indicates that the robustness of our data construction process which minimizes dependence on super powerful math expert models for generating synthetic data. 

Second, we analyze the effectiveness of different math-specific samples for math SFT. To study this, we compare AceMath-Instruct trained with 1.6 million math SFT samples (details in \S\ref{sec:train_strategy_math_sft}) to models trained using all available math SFT samples (2.3 million) or only cross-checked high-quality samples (800K). We find that simply increasing the quantity of data or exclusively using high-quality samples does not yield better outcomes. Instead, combining cross-checked high-quality data with additional samples that include a diverse range of math questions produces superior results.

Third, we study the impact of conducting general SFT before transitioning to math SFT. To explore this, we skip the general SFT step, and conduct math SFT directly using all math-specific samples. We observe that skipping general SFT typically results in an average score drop of approximately 1\%, even when using a math-base model (e.g., Qwen2.5-Math-72B-Base) as the backbone. The results highlight the effectiveness of conducting general SFT prior to math SFT.

\subsubsection{Ablation Studies on Synthetic Data}
\label{sec:ablation_synthetic_data}
As shown in Table~\ref{table:sft_math_ablation_synthetic}, we study how synthetic math SFT data affects the results. We compare AceMath-Instruct against two scenarios: one where all one million synthetic data samples are removed and another where an additional 500K low-quality synthetic data are included for training (e.g., lengthy prompts and one type of \textit{in-depth evolution} that adds constraints). Details of the synthetic math SFT data can be found in \S\ref{sec:math_sft}.
In both scenarios, we observe a decline in results, underscoring the importance of not only generating synthetic data but also carefully selecting it for training.
Effectively leveraging appropriate synthetic data proves essential for achieving optimal performance.

\section{Reward Model Training}
We train a math reward model for AceMath-Instruct, aiming to select more accurate solutions and better reasoning paths. 
To ensure broad applicability across a variety of language models, we curate a diverse training dataset. The following sections detail our training methodology, evaluation protocols, and empirical results.

\label{sec:reward_modeling}
\subsection{Reward Training Data Synthesis}
\label{sec:reward_modeling_data_synthesis}
\subsubsection{Initial Dataset Construction}
We utilize a portion of the math SFT dataset (350K) from \S\ref{sec:math_sft} to use the prompts and the answers generated by gpt-4o-mini~\citep{gpt4omini} as reference labels.
To capture the diversity of model-generated reasoning steps and potential different kinds of reasoning mistakes, we sample four model responses per LLM from a set of 14 LLMs, including Llama2-7b-chat~\citep{touvron2023llama}, Llama3.1-8/70B-Instruct~\citep{dubey2024llama}, DeepSeek-math-7b-instruct~\citep{shao2024deepseekmath}, Mistral-7B/Mathstral-7B~\citep{jiang2023mistral}, Gemma-2/27b-it~\citep{team2024gemma}, and Qwen2/2.5-1.5/7/72B-Instruct~\citep{yang2024qwen2_5_math}.
We then annotate the model solutions as correct or incorrect by comparing them against the referenced labels using the Qwen-math evaluation toolkit.~\footnote{\url{https://github.com/QwenLM/Qwen2.5-Math/tree/main/evaluation}} This process initializes a pool of correct and incorrect candidate responses for each problem, which we treat as positive and negative samples that can be further sampled to create paired responses for training.

\subsubsection{Response Scoring and Selection}
\label{sec:reward_modeling_data_sampling}
Mathematical problem answers encompass a wide range of formats with diverse representations (\textit{e.g.,} [\textit{\textbackslash frac\{1\}\{2\}, 1/2, 0.5}] and [\textit{1e-5, 1$\times$10\textasciicircum\{-5\}}]), and heuristic math evaluation toolkits using SymPy and latex2sympy2 may inevitably result in false negative candidates (\textit{i.e.,} correct answers annotated as incorrect). Such examples in the negative candidates could introduce noise and adversely affect model training.
Therefore, instead of randomly sample responses from all candidates, we rank the candidates and apply a score-sorted sampling strategy.
Specifically, we use the math reward model Qwen2.5-Math-RM-72B to rank positive and negative candidates for each problem based on their scores.
We then randomly sample from top-$k$ positive and bottom-$k$ negative candidates, with $k$ set to 14 based on preliminary experiments.
Compared to random sampling from all candidates, our ablation study in Table~\ref{tab:rm_ablation} demonstrates the benefits of the score-sorted sampling strategy.
In conclusion, we sample a total of six response candidates (positive + negative) for each problem, ensuring a balanced number of positive and negative responses, and filter out problems where all responses are either correct or incorrect. 

\subsubsection{Addressing Stylistic Biases}
LLMs can generate different styles of chain-of-thought reasoning paths when prompted in the zero-shot setting or with few-shot examples~\citep{wei2022chain}. We observe significant shorter and simple reasoning paths in model outputs for datasets such as MMLU~\citep{hendrycks2021measuring} as the model follows the simple 5-shot examples provided in the instruction. 
To improve reward model performance on such output styles, we create training data using the few-shot prompting approach to generate simple and short reasoning paths for 2,000 multiple-choice problems.
In addition, as our ultimate goal is to develop a reward model for the AceMath-Instruct model family, we sample a set of 30,000 problems and use AceMath-(1.5/7/72B)-Instruct checkpoints to generated responses to create positive and negative pairs for training. In conclusion, our final training dataset consists of 356K problems, each paired with a total of six responses ($k$ positive and $6-k$ negative).

\begin{table*}[htp]
  \centering
  \resizebox{\linewidth}{!}{
    \begin{tabular}{lcccccccc}
    \toprule
    Model & GSM8K & MATH500  & \makecell{Minerva\\Math} & \makecell{GaoKao\\2023 En} & \makecell{Olympiad\\Bench} & \makecell{College\\Math} & \makecell{MMLU\\STEM} & \makecell{Avg.} \\
\midrule
        majority@8	& 96.22	& 83.11	& 41.20	& 68.21	& 42.69	& 45.01	& 78.21	& 64.95 \\
        
{\small Internlm2-7b-reward} & 95.26 & 78.96 & 36.25 & 67.51 & 40.49 & 43.88 & 75.42 & 62.54 \\
{\small Internlm2-20b-reward} & 95.10 & 76.53 & 37.69 & 66.63 & 40.12	& 42.57	& 70.60 &	61.32\\
{\small Skywork-Reward-Llama-3.1-8B-v0.2} & 95.64 & 74.16 & 39.11 & 67.16 & 39.10 & 44.58 & 76.52 & 62.32 \\
{\small Skywork-Reward-Gemma-2-27B-v0.2} & 95.94 & 74.90 & 39.37 & 66.96 & 39.07 & 45.46 & 78.20 & 62.84 \\
{\small Skywork-o1-Open-PRM-Qwen-2.5-7B} & 96.92	&86.64	&41.00	&72.34	&46.50&	46.30&	74.01 & 66.24 \\
Qwen2.5-Math-RM-72B & 96.61	& 86.63	& 43.60	& 73.62	& 47.21	& \textbf{47.29}	& 84.24	& 68.46\\
\addlinespace[1.5pt]
\hdashline
\addlinespace[2.5pt]
AceMath-7B-RM (Ours) & 96.66&	85.47&	41.96&	73.82&	46.81&	46.37&	80.78&	67.41\\
AceMath-72B-RM (Ours)& \textbf{97.23}&	\textbf{86.72}&	\textbf{45.06}&	\textbf{74.69}&	\textbf{49.23}&	46.79&	\textbf{87.01}&	\textbf{69.53}\\
\midrule
\color{gray!80} pass@8 (Oracle) &	\color{gray!80} 98.86	& \color{gray!80} 91.84	& \color{gray!80} 56.18	& \color{gray!80} 82.09	& \color{gray!80} 59.00	& \color{gray!80} 56.38	& \color{gray!80} 96.15	& \color{gray!80} 77.21\\
    \bottomrule
    \end{tabular}%

    }
     \caption{\textbf{Reward model evaluation on AceMath-RewardBench}. The average results (rm@8) of reward models on math benchmarks, randomly sample 8 responses from 64 candidates with 100 random seeds. Response candidates are generated from a pool of 8 LLMs (Qwen\{2/2.5\}-Math-\{7/72\}B-Instruct, Llama-3.1-\{8/70\}B-Instruct, Mathtral-7B-v0.1, deepseek-math-7b-instruct).}
\label{tab:rm_results}%

\end{table*}%
\subsection{Reward Training Strategy}
Our reward model architecture adopts a outcome reward approach, which introduces a linear layer at the top of the language model to project the last token representation into a scalar value. We initialize the backbone of the reward model using a supervised fine-tuned model (\textit{i.e.,} AceMath-Instruct).
Following the training objective established in Qwen2.5-Math~\citep{yang2024qwen2_5_math}, we construct problem-response pairs with $k$ positive (correct) candidates and $6-k$ negative (incorrect) candidates. 
We compute the list-wise Bradley-Terry loss~\citep{bt-1952}, which demonstrates computational efficiency compared to pair-wise approaches as shown in Table~\ref{tab:rm_ablation}.

{\small
\begin{align*}
&\mathcal{L}_{\text{rm}}(\theta) =
\\
& -\frac{1}{k\cdot(6-k)}\mathbb{E}_{(x, y_\text{pos}, y_\text{neg})} \Big[
\log \big(\sigma ( 
r_\theta(x, y_\text{pos}) - r_\theta(x, y_\text{neg}))
\big)
\Big]
\end{align*}
}

Here, $r_\theta(x, y)$ represents the output score of the reward model $r_\theta$, where $x$ denotes the problem and $y$ represents the response candidate. The loss function is designed to optimize the model's ability to discriminate between correct and incorrect responses by maximizing the margin between positive and negative candidate scores.

\subsection{Reward Evaluation Benchmarks}
\subsubsection{AceMath-RewardBench}
Existing math reward benchmarks lack diversity, both in the types of candidate solutions and the range of difficulty levels in the math questions.
To address this, we construct a math reward model evaluation benchmark, AceMath-RewardBench, which contains 7 datasets and use 8 different LLMs to generate solutions for robust evaluation.
The benchmark use the best-of-N (BoN or rm@$n$) metric, a methodology extensively used in literature~\citep{cobbe2021training,lightman2023let,yang2024qwen2_5_math}. 
The primary objective of the reward model is to select the highest reward scored model response from a candidate set of $n$ and calculate the corresponding problem-solving rate for each math benchmark (7 datasets) used in \S\ref{sec:math_sft_benchmark}.
We adopt rm@8 metric following the Qwen2.5-Math evaluation protocol, optimizing computational efficiency during the inference stage. 
To ensure robust and statistically reliable benchmark performance, we implement two design principles: 1) diverse model distribution: we sample 8 responses from each model in a set of mathematical and general-purpose LLMs (\textit{i.e.,} Qwen2.5-Math-7/72B-Instruct~\citep{yang2024qwen2_5_math}, Qwen2-Math-7/72B-Instruct~\citep{yang2024qwen2}, Llama-3.1-8/70B-Instruct~\citep{dubey2024llama}, DeepSeek-Math-7B-Instruct~\citep{shao2024deepseekmath}, Mathtral-7B-v0.1~\citep{jiang2023mistral}), mitigating potential model-specific style biases; 2) we compute accuracy metrics by averaging results across 100 random seeds, reducing result variance and enhancing reproducibility.
    
In total, each problem in the benchmark contains a total of 64 candidate responses from 8 LLMs.
We then randomly sample 8 responses from these 64 candidates, compute the rm@8 result, and average the final accuracy over 100 random seeds.
Different from the Math SFT evaluation, we use the MATH500~\citep{lightman2023let}, a subset of 500 problems sampled from the MATH dataset~\citep{hendrycksmath2021} following the prior work such as PRM800K~\citep{lightman2023let} and RewardBench~\citep{lambert2024rewardbench}.

\subsubsection{RewardBench (MATH500) and RewardMath} 
Apart from our own benchmarks, we also evaluate on RewardBench~\citep{lambert2024rewardbench} (MATH500) and RewardMath~\citep{kim2024evaluating} to report the accuracy of selecting the correct solution from a list of candidates for each problem in MATH500~\citep{lightman2023let}. The primary difference between these two benchmarks lies in the candidate sets: RewardBench uses one correct (human-written) solution and one incorrect candidate (generated by GPT-4), while RewardMath uses one correct (a GPT-4 rewrite) and nine incorrect candidates (generated by models). \citet{kim2024evaluating} highlight a significant distributional shift between human-written solutions and machine-generated ones as the prior ones are typically shorter, more concise, and contain fewer details. 
This difference in style and content may partially explain why saturated accuracy exceeding 95\% on the RewardBench. To address this limitation and better assess the robustness of reward models, they propose RewardMath, which introduces a more challenging evaluation setup and show that most reward models struggle significantly with this new benchmark, achieving accuracies of only around 30\% or lower.

\subsection{Experiments of Reward models}
\subsubsection{Hyperparameters} We use the AceMath-7B/72B-Instruct model as the backbone to train the outcome reward model: AceMath-RM-7/72B. The model is trained using AdamW~\citep{kingma2014adam,loshchilov2017decoupled} for 2 epochs with a learning rate of \{5e-6, 2e-6\}, using a cosine learning rate scheduler and an effective batch size of 256. Training is conducted on 8 H100 GPUs for the 7B model and 256 H100 GPUs for the 72B model.

\subsubsection{Baselines}
For mathematical reward modeling, we compare with current state-of-the-art outcome reward model Qwen2.5-Math-RM-72B~\citep{yang2024qwen2_5_math} and a process reward model Skywork-o1-Open-PRM-Qwen-2.5-7B~\citep{skyworkopeno12024}.
We also include majority@8 (majority voting) baseline and the pass@8 (any one of the 8 is correct) as an oracle reward model to measure the upper bound of this benchmark. 
Additionally, we incorporate general reward models top-ranked on RewardBench, including Skywork-Reward~\citep{liu2024skywork} and Internlm2-reward~\citep{cai2024internlm2}. It is noteworthy that while these models are not exclusively trained for mathematical domains, a substantial portion of their training data encompasses mathematical content. For instance, Skywork-Reward~\citep{liu2024skywork} uses 50\% math data for training.

\subsubsection{Results on AceMath-RewardBench} 
In Table~\ref{tab:rm_results}, we show that our AceMath-72B-RM achieves the state-of-the-art rm@8 accuracy on average of AceMath-RewardBench, outperforming the Qwen2.5-Math-RM-72B by 1\% absolute (69.53 vs 68.46) and on 6 out of 7 datasets. 
We show the 7B variant achieves 67.41 accuracy on average and demonstrates the benefits of model size scaling from 7B to 72B, especially on datasets require college-level STEM knowledge such as Minerva Math (41.96 $\to$ 45.06) and MMLU STEM (80.78 $\to$ 87.01). Comparing to other reward model baselines, the 7B outperform Internlm2 and Skywork-Reward by a large margin as our benchmark reveal these reward model even underperform the majority voting baseline. Nevertherless, we note that there remains considerable room for improvement as indicated by the gap between the reward model and pass@8 oracle accuracy.

\begin{table}[t]
  \centering
  \resizebox{\linewidth}{!}{
    \begin{tabular}{lcccccccc}
    \toprule
    Model & \makecell{RewardBench \\ {\scriptsize MATH500}} & \makecell{RewardMath \\ {\scriptsize MATH500}} \\
\midrule        
\textit{Random} & 50.00 & 10.00 \\
\midrule
\color{gray} \textit{LLM-as-a-Judge}\\
Claude-3.5-Sonnet$^\dagger$ & 70.70 & 15.32\\
GPT-4o-2024-05-13$^\dagger$ & 72.50 & 25.98\\
\midrule
\color{gray} \textit{Classifier-based}\\
Math-Shepherd-Mistral-7B$^\dagger$ & 94.41 & 17.18\\
ArmoRM-Llama3-8B-v0.1$^\dagger$ & \textbf{98.70} & 20.50 \\
Skywork-Reward-Llama-3.1-8B$^\dagger$ & 96.87	& 22.15	 \\
Internlm2-20b-reward$^\dagger$ & 95.10	& 33.95	 \\
Internlm2-7b-reward$^\dagger$ & 94.90	& 37.27	 \\
{Skywork-o1-Open-PRM-7B} & 78.52 & 51.34\\
Qwen2.5-Math-RM-72B & 95.97	& \underline{68.53}	\\
\addlinespace[2.5pt]
\hdashline
\addlinespace[2.5pt]
AceMath-7B-RM (Ours) & 92.62	& 57.76	\\
AceMath-72B-RM (Ours) & \underline{97.09} & \textbf{68.94}\\
    \bottomrule
    \end{tabular}%
    }
    \vspace{-.3cm}
     \caption{The accuracy of reward models on RewardBench (MATH500)~\citep{lambert2024rewardbench} and RewardMATH~\citep{kim2024evaluating}. $\dagger$: Results are copied from RewardMATH. \textbf{Bold}: top-1. \underline{Underline}: top-2 accuracy.}
\label{tab:rm_bench}%

\end{table}%

\subsubsection{Results on RewardBench and RewardMath} 
In Table~\ref{tab:rm_bench}, we demonstrate that our AceMath-72B-RM achieves state-of-the-art accuracy on RewardMATH. 
While many reward models (e.g., ArmoRM~\citep{ArmoRM}, Internlm2) achieve 95\%+ accuracy on the RewardBench MATH500 split, their accuracy drops significantly on RewardMATH, ranging from only 20\% to 37\%. 
We found Skywork-PRM model performs much better on RewardMATH (51.34) but worse on RewardBench (78.5). This may be due to the lack of reasoning steps typically found in human-written solutions, and as a result, our AceMath-7B-RM outperforms it on both benchmarks. In conclusion, these evaluation results highlight the benefits of training on diverse, model-generated solutions to mitigate, though not entirely eliminate, out-of-distribution generalization challenges.

\begin{table}[t]
  \centering
  \resizebox{\linewidth}{!}{
    \begin{tabular}{lc}
    \toprule
    Model & AceMath-RewardBench\\
\midrule
AceMath-7B-RM & \textbf{67.41}\\
\quad $\triangleright$ Backbone: {\footnotesize Qwen2.5-Math-7B-Instruct} & 66.93\\
\quad $\triangleright$ Data: Random sampling & 67.07\\
\quad $\triangleright$ Loss: Pairwise BT & 67.33 \\
\quad $\triangleright$ Loss: Cross-entropy Classification & 66.93 \\
\quad $\triangleright$ Loss: MSE Regression & 66.79\\
\addlinespace[2.5pt]
\hdashline
\addlinespace[2.5pt]
AceMath-72B-RM & \textbf{69.53} \\
\quad $\triangleright$ Backbone: {\footnotesize Qwen2.5-Math-72B-Instruct} & 69.09\\
\quad $\triangleright$ Loss: Cross-entropy Classification & 68.66 \\
    \bottomrule
    \end{tabular}%
    }
     \caption{Ablation study of AceMath-7/72B-RM on AceMath-RewardBench (Backbone: AceMath-7/72B-Instruct; Data: reward score-sorted sampling; Loss: listwise Bradley-Terry.}
\label{tab:rm_ablation}%

\end{table}%

\subsubsection{Ablation studies}
In Table~\ref{tab:rm_ablation}, we conduct ablation studies on the model backbone, data sampling method, and different loss functions used to train the reward model. 
First, we found that using AceMath-7B-Instruct as the backbone model for training the model consistently outperforms Qwen2.5-Math-7B-Instruct on average of 7 datasets, with a similar performance gap observed at the 72B scale. 
Secondly, we observed that employing reward score-sorted sampling (\S\ref{sec:reward_modeling_data_sampling}) during the data construction process improves average accuracy compared to random sampling. This highlights the benefits of filtering out noisy labels when heuristic evaluation toolkits produce false negative errors. 
Lastly, we experimented with different loss functions. We found that using pairwise Bradley-Terry loss achieves comparable accuracy to the listwise loss approach, however requiring 3.7$\times$ more training time using 8 H100 GPUs. 
Additionally, training a classifier using cross-entropy loss or a regression model using mean squared error (MSE) loss both resulted in lower accuracy. A similar performance gap was also observed at the 72B scale for the cross-entropy classification approach. 
Since the data is constructed for the listwise BT approach, where each problem consists of six responses, this also leads to 3.8 times more compute hours on 8 GPUs.

\begin{figure}[t]
    \centering
    \includegraphics[width=0.99\linewidth]{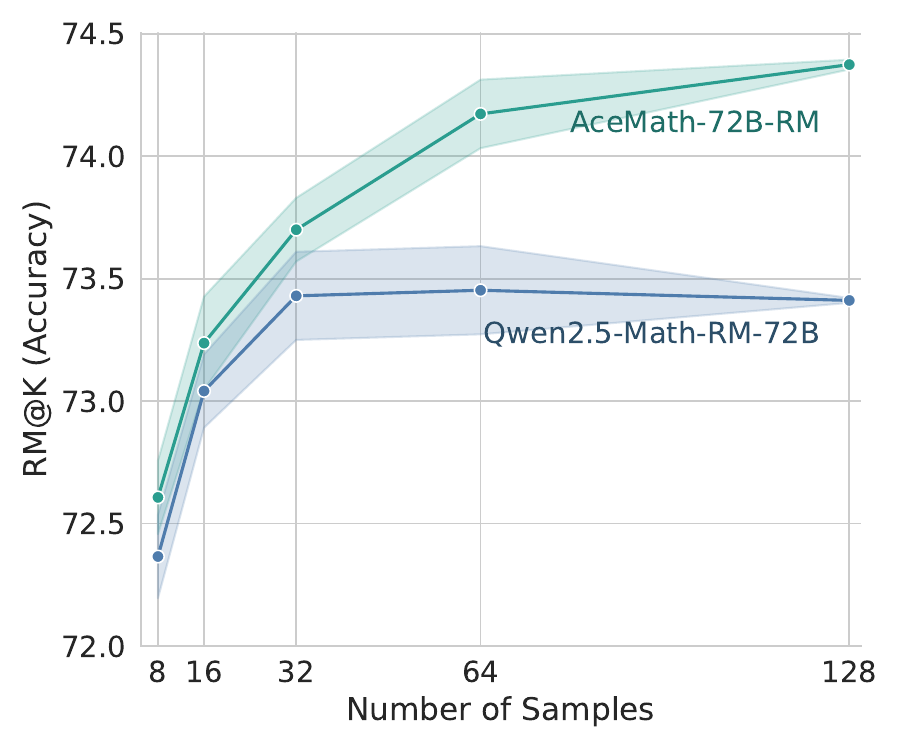}
    \vspace{-1em}
    \caption{rm@$k$ evaluation on average accuracy of 7 datasets for AceMath-7B-Instruct.}
    \label{fig:rm_scale}
\end{figure}
\begin{figure*}[ht]
    \centering
    \includegraphics[width=0.99\linewidth]{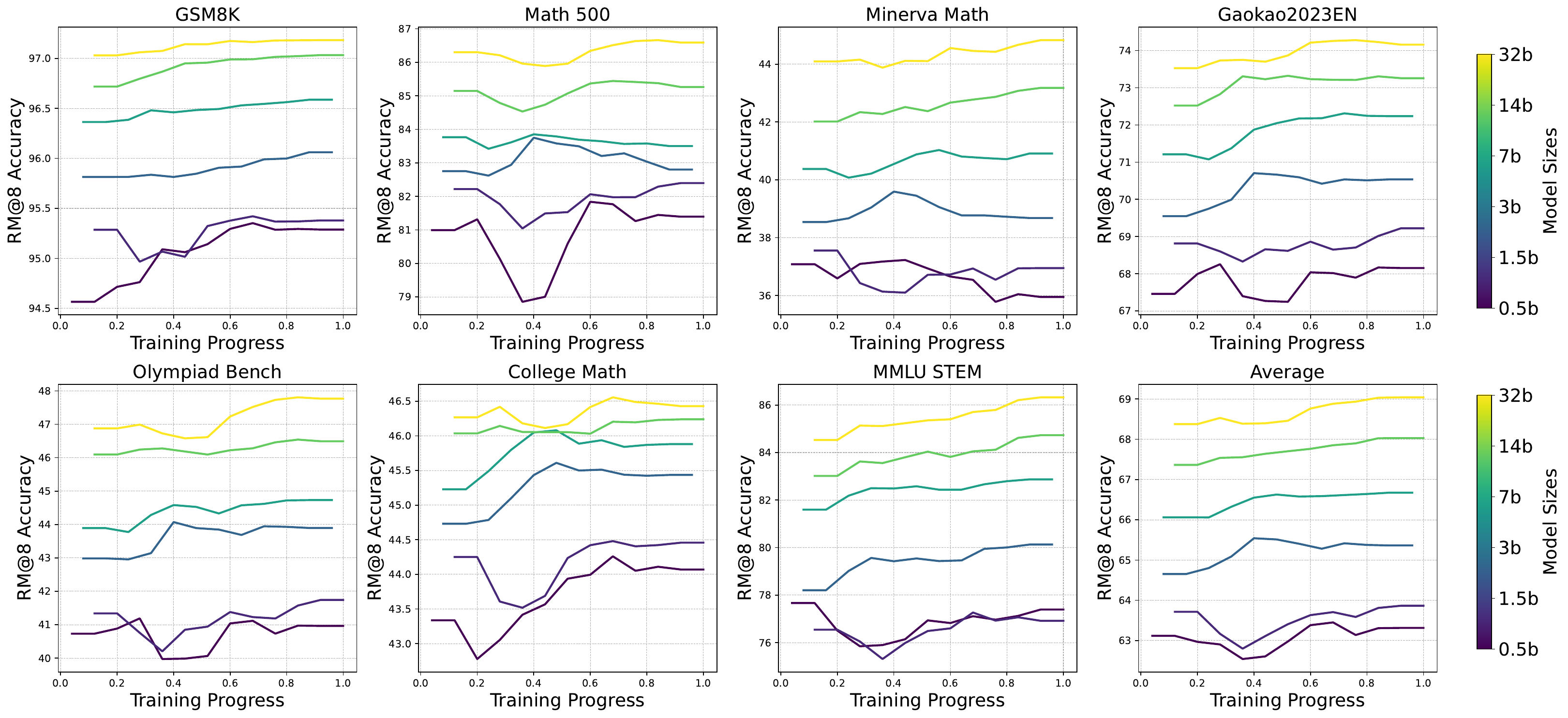}
    \vspace{-1em}
    \caption{Learning curves for reward model training. All models are trained from Qwen2.5-Instruct family.}
    \label{fig:rm_learning}
\end{figure*}
\subsubsection{Results on rm@$k$}
In Figure~\ref{fig:rm_scale}, we present a comparison between AceMath-72B-RM and Qwen2.5-Math-RM-72B on rm@$k$ ($k=8,16,32,64,128$) across the seven datasets listed in Table~\ref{tab:rm_results}, using samples generated by AceMath-7B-Instruct. We report the average accuracy across these seven datasets, each with 10 different random seeds.

First, we find that using AceMath-72B-RM to score the outputs from AceMath-7B-Instruct consistently improves the average accuracy, increasing from 72.6 to 74.4 as $k$ rises from 8 to 128.
Second, we observe that AceMath-RM consistently outperforms Qwen2.5-Math-RM in scoring outputs generated from AceMath-7B-Instruct, and this improvement becomes larger as $k$ increases.

Furthermore, we compare the performance of AceMath-72B-RM paired with AceMath-Instruct to Qwen2.5-Math-RM-72B paired with Qwen2.5-Math-Instruct. As shown in Figure~\ref{fig:main_ressults}, the AceMath combination consistently outperforms its Qwen2.5 counterpart in terms of rm@8, on average, for both the 7B and 72B models.
Remarkably, we find that the our AceMath-7B model even outperforms the Qwen2.5-Math-72B in rm@8, showing the potential of a smaller model when paired with a carefully designed reward model.

\subsubsection{Learning curves of reward model training}
In Figure~\ref{fig:rm_learning}, we aim to understand how reward modeling accuracy improves as we increase model size and use additional data. We find distinct patterns in the interplay between model size and data scaling. In general on simpler dataset such as GSM8K, all model sizes (ranging from 0.5B to 32B parameters) exhibit steady improvements as training proceeds, with larger models achieving higher accuracy. In contrast, on the more challenging datasets, which requires college-level knowledge, such as Minerva Math, MMLU STEM, and OlympiadBench, model size emerges as a critical factor: smaller models (\textit{e.g.,} 0.5B, 1.5B) show negligible improvement despite increased data, whereas larger models (\textit{e.g.,} 14B, 32B) achieve better accuracy gains. These results suggest that increasing model size provides the greatest benefit, whereas the advantages of increasing data appear less pronounced.
Our experiments use Qwen2.5-Instruct~\citep{yang2024qwen2} model family. instead of Qwen2.5-Math-Instruct, as it provides a more comprehensive set of models with different sizes. All models are trained for one epoch only.

\section{Conclusion}
\label{sec:conclusion}

In this work, we present AceMath, a series of frontier-class math instruct and reward models. 
We demonstrate that our AceMath-7B-Instruct significantly surpasses the previous best-in-class Qwen2.5-Math-7B-Instruct across comprehensive math reasoning benchmarks, and it performs slightly worse than a 10$\times$ larger Qwen2.5-Math-72-Instruct (67.2 vs. 68.2). Remarkably, our AceMath-72B-Instruct greatly outperforms Qwen2.5-Math-72-Instruct, GPT-4o and Claude-3.5 Sonnet.
Additionally, we construct AceMath-RewardBench, a comprehensive benchmark designed to evaluate math reward models across a diverse range of datasets and difficulty levels. We show that our AceMath-72B-RM consistently outperforms state-of-the-art reward models, including Qwen2.5-Math-RM-72B and Skywork-o1-Open-PRM-Qwen-2.5-7B, on various math reward benchmarks.
Furthermore, when combining AceMath-72B-Instruct with AceMath-72B-RM, we achieve the highest average rm@8 score across the math reasoning benchmarks.
To advance open research in the field, we will open source the model weights for both AceMath-Instruct and AceMath-RM, along with the complete training data used throughout their development.

\bibliography{paper}
\bibliographystyle{icml2024}

\newpage
\appendix
\onecolumn

\section{AIME 2024 \& AMC 2023 Results}
\label{appendix:amc_aime_results}
\begin{table}[h]
\centering
\resizebox{0.45\linewidth}{!}{
\begin{tabular}{lcc}
\toprule
Models     &  AIME 2024  & AMC 2023 \\ \midrule

Llama-3.1-405B-Instruct  & 5/30  & 20/40 \\

Claude 3.5 Sonnet (2024-1022) & 4/30  & 21/40 \\

OpenMath2-Llama3.1-8B & 3/30 &  16/40 \\

OpenMath2-Llama3.1-70B &  4/30 &  20/40\\

Qwen2.5-Math-1.5B-Instruct & 3/30  & 24/40 \\

Qwen2.5-Math-7B-Instruct & 5/30 & 25/40  \\

Qwen2.5-Math-72B-Instruct & \textbf{9/30} & \textbf{28/40} \\ \midrule

AceMath-1.5B-Instruct &  4/30  & 25/40 \\
AceMath-7B-Instruct &  6/30  & 26/40  \\
AceMath-72B-Instruct & 6/30  &  \textbf{28/40} \\ \bottomrule

\end{tabular}
}
\caption{Greedy decoding results of AceMath-Instruct on AIME 2024 and AMC 2023.}
\label{table:sft_math_aime_amc_results}
\end{table}

Table~\ref{table:sft_math_aime_amc_results} shows the greedy decoding results on AIME 2024 and AMC 2023. We find that the AceMath-1.5B/7B-Instruct models slightly outperform Qwen2.5-Math-1.5B/7B-Instruct on both datasets, while AceMath-72B-Instruct falls short of Qwen2.5-Math-72B-Instruct's performance on AIME 2024. 
Given that AIME 2024 contains challenging math problems comparable to pre-Olympiad levels, these results indicate that there is room to improve AceMath-Instruct to better address varying levels of mathematical difficulty.

\section{AceMath-Instruct Using Different Backbone Models}
\label{appendix:full_base_mathbase_backbone}
\begin{table}[h]
\centering
\resizebox{0.99\linewidth}{!}{
\begin{tabular}{lcccccccc}
\toprule
Models                                 & GSM8K & MATH  & \begin{tabular}[c]{@{}c@{}}Minerva\\ Math\end{tabular} & \begin{tabular}[c]{@{}c@{}}GaoKao\\ 2023 En\end{tabular} & \begin{tabular}[c]{@{}c@{}}Olympiad\\ Bench\end{tabular} & \begin{tabular}[c]{@{}c@{}}College\\ Math\end{tabular} & \begin{tabular}[c]{@{}c@{}}MMLU\\ STEM\end{tabular} & Avg.  \\ \midrule
DeepSeek-Math-7B-RL      & 88.20 & 52.40 & 20.60 & 43.60   & 19.00   & 37.50   & 64.80    & 46.59 \\
AceMath-Instruct (backbone: DeepSeek-Coder-7B-Base) & 83.85 & 59.72 & 29.78 & 53.51 & 24.59 & 44.64 & 55.95 & 50.29 \\
AceMath-Instruct (backbone: DeepSeek-Math-7B-Base)  & 85.06 & 66.86 & 40.07  & 56.62   & 29.63   & 48.94    & 65.53     & \textbf{56.10} \\ \midrule

Llama-3.1-8B-Instruct   & 84.50 & 51.90 & 21.70    & 38.40    & 15.40      & 33.80    & 60.50          & 43.74 \\
OpenMath2-Llama3.1-8B   & 91.70 & 67.80 & 16.91 & 53.76 & 28.00 & 46.13 & 46.02 & 50.08 \\
AceMath-Instruct (backbone: Llama3.1-8B-Base)       & 91.51 & 69.06 & 31.99  & 59.74  & 32.00  & 49.08  & 67.94  & \textbf{57.33} \\ \midrule

Qwen2.5-Math-1.5B-Instruct      & 84.80 & 75.80 & 29.40   & 65.50  & 38.10   & 47.70    & 57.50   & 56.97 \\
AceMath-Instruct (backbone: Qwen2.5-1.5B-Base)  & 80.89 & 64.59 & 30.51 & 53.25 & 27.11 & 47.80 & 58.62 & 51.82  \\
AceMath-Instruct (backbone: Qwen2.5-Math-1.5B-Base) & 86.95 & 76.84 & 41.54  & 64.42   & 33.78    & 54.36  & 62.04    & \textbf{59.99} \\ \midrule

Qwen2.5-Math-7B-Instruct       & 95.20 & 83.60 & 37.10  & 66.80 & 41.60  & 46.80   & 71.90  & 63.29 \\
AceMath-Instruct (backbone: Qwen2.5-7B-Base) & 93.56 & 77.10 & 43.38 & 65.19 & 37.78 & 54.90 & 77.41 & 64.19  \\
AceMath-Instruct (backbone: Qwen2.5-Math-7B-Base)   & 93.71 & 83.14 & 51.11  & 68.05   & 42.22   & 56.64  & 75.32  & \textbf{67.17} \\ \midrule

Qwen2.5-Math-72B-Instruct     & 95.90 & 85.90 & 44.10   & 71.90   & 49.00   & 49.50   & 80.80   & 68.16 \\ 
AceMath-Instruct (backbone: Qwen2.5-72B-Base) & 95.99 & 85.06 & 54.04 & 73.25 & 46.96 & 57.10 & 85.48 & 71.13  \\
AceMath-Instruct (backbone: Qwen2.5-Math-72B-Base)  & 96.44 & 86.10 & 56.99   & 72.21  & 48.44   & 57.24   & 85.44    & \textbf{71.84} \\ \bottomrule
\end{tabular}
}
\caption{Greedy decoding results of AceMath-Instruct across different backbone models.}
\label{table:sft_math_mathbase_versus_base_backbones_full}
\end{table}

Table~\ref{table:sft_math_mathbase_versus_base_backbones_full} shows the full results of AceMath-Instruct using various models as backbone models. Additionally, we include the results for Llama3.1-8B-Base as the backbone model and compare our model to OpenMath2-Llama3.1-8B~\citep{toshniwal2024openmathinstruct} that also uses Llama3.1-8B-Base as its backbone model. We find that except for our 1.5B model based on Qwen2.5-1.5B-Base, all our models, including those built on base models, outperform their respective strong baselines, often by a significant margin.

\section{Synthetic Prompt Generation for Math SFT}
\label{appendix:synthetic_prompt_generation}
In this section, we describe the prompt we provided to GPT-4o-mini (2024-0718) for generating synthetic prompts tailored to math SFT. We utilize the in-breath evolution and in-depth evolution prompts inspired from~\citet{xu2024wizardlm}.

\subsection{In-Breath Evolution}
We use the following prompt to generate more diverse math questions.
\begin{verbatim}
You are a good math question creator.

Your objective is to draw inspiration from the #Given MATH Question# to create a 
brand new math question. This new math question should be distinctly different from 
the #Given MATH Question# and be even more unique.

The length and difficulty level of the #Created MATH Question# should be similar to 
those of the #Given MATH Question#.

The #Created MATH Question# must be solvable and understandable by humans.

#Given MATH Question#:
{given_math_question}

#Created MATH Question#:

\end{verbatim}

\subsection{In-Depth Evolution}
We use the following prompt to generate more challenging math questions.

\begin{verbatim}
You are a good math question creator.

Your objective is to draw inspiration from the #Given MATH Question# to create a 
brand new math question. This new math question should be more complex and 
challenging than the #Given MATH Question#.

The #Created MATH Question# must be solvable and understandable by humans.

#Given MATH Question#:
{given_math_question}

#Created MATH Question#:

\end{verbatim}

Moreover, we find that the following prompt that requires to add constraints to the given prompt could result in unsolvable or overly challenging math questions. This, in turn, can lead to incorrect answers being included in the training data, ultimately degrading model performance.

\begin{verbatim}
You are a good math question creator.

Your objective is to rewrite the #Given MATH Question# into a brand new but more 
complex version. You can complicate the #Given MATH Question# by introducing 
additional constraints and requirements.

The #Created MATH Question# must be solvable and understandable by humans.

#Given MATH Question#:
{given_math_question}

#Created MATH Question#:

\end{verbatim}

\section{Chain-of-Thought Reasoning Examples}
\label{appendix:cot_examples}
Below, we present several Chain-of-Thought reasoning examples generated by AceMath-72B-Instruct.

\rule{\textwidth}{1pt}
\textsc{\textbf{Sample 1}} \\ \\
\textit{\textbf{Question:}} \\
For each positive integer $N$, let $P(N)$ denote the product of the digits of $N$. For example, $P(8)=8$, $P(451)=20$, and $P(2023)=0$. Compute the least positive integer $n$ such that $P(n+23)=P(n)+23$. 
\\
\rule{\textwidth}{0.5pt}

\textit{\textbf{AceMath-72B-Instruct:}} \\
To find the least positive integer \( n \) such that \( P(n+23) = P(n) + 23 \), we need to analyze the properties of the function \( P \), which gives the product of the digits of a number. Let's start by considering the possible values of \( n \) and their corresponding values of \( P(n) \) and \( P(n+23) \).

First, note that if \( n \) has a digit of 0, then \( P(n) = 0 \). In this case, \( P(n+23) = 23 \), which means \( n+23 \) must have digits whose product is 23. Since 23 is a prime number, the only way to achieve this is if \( n+23 \) is either 23 or a number with 2 and 3 as its only digits. However, \( n \) must be a positive integer, so \( n+23 \) cannot be 23. Therefore, \( n+23 \) must be a number like 230, 203, 320, 302, etc. The smallest such number is 230, so \( n = 230 - 23 = 207 \). Let's verify this solution:

- \( P(207) = 2 \cdot 0 \cdot 7 = 0 \)

- \( P(230) = 2 \cdot 3 \cdot 0 = 0 \)

This does not satisfy the equation \( P(n+23) = P(n) + 23 \). Therefore, we need to consider other possibilities. Next, let's consider the case where \( n \) does not have a digit of 0. We need to find \( n \) such that the product of the digits of \( n+23 \) is 23 more than the product of the digits of \( n \). Let's try some small values of \( n \):

- If \( n = 1 \), then \( P(1) = 1 \) and \( P(24) = 2 \cdot 4 = 8 \). \( 8 \neq 1 + 23 \).

- If \( n = 2 \), then \( P(2) = 2 \) and \( P(25) = 2 \cdot 5 = 10 \). \( 10 \neq 2 + 23 \).

- If \( n = 3 \), then \( P(3) = 3 \) and \( P(26) = 2 \cdot 6 = 12 \). \( 12 \neq 3 + 23 \).

- If \( n = 4 \), then \( P(4) = 4 \) and \( P(27) = 2 \cdot 7 = 14 \). \( 14 \neq 4 + 23 \).

- If \( n = 5 \), then \( P(5) = 5 \) and \( P(28) = 2 \cdot 8 = 16 \). \( 16 \neq 5 + 23 \).

- If \( n = 6 \), then \( P(6) = 6 \) and \( P(29) = 2 \cdot 9 = 18 \). \( 18 \neq 6 + 23 \).

- If \( n = 7 \), then \( P(7) = 7 \) and \( P(30) = 3 \cdot 0 = 0 \). \( 0 \neq 7 + 23 \).

- If \( n = 8 \), then \( P(8) = 8 \) and \( P(31) = 3 \cdot 1 = 3 \). \( 3 \neq 8 + 23 \).

- If \( n = 9 \), then \( P(9) = 9 \) and \( P(32) = 3 \cdot 2 = 6 \). \( 6 \neq 9 + 23 \).

- If \( n = 10 \), then \( P(10) = 1 \cdot 0 = 0 \) and \( P(33) = 3 \cdot 3 = 9 \). \( 9 \neq 0 + 23 \).

- If \( n = 11 \), then \( P(11) = 1 \cdot 1 = 1 \) and \( P(34) = 3 \cdot 4 = 12 \). \( 12 \neq 1 + 23 \).

- If \( n = 12 \), then \( P(12) = 1 \cdot 2 = 2 \) and \( P(35) = 3 \cdot 5 = 15 \). \( 15 \neq 2 + 23 \).

- If \( n = 13 \), then \( P(13) = 1 \cdot 3 = 3 \) and \( P(36) = 3 \cdot 6 = 18 \). \( 18 \neq 3 + 23 \).

- If \( n = 14 \), then \( P(14) = 1 \cdot 4 = 4 \) and \( P(37) = 3 \cdot 7 = 21 \). \( 21 = 4 + 17 \).

- If \( n = 15 \), then \( P(15) = 1 \cdot 5 = 5 \) and \( P(38) = 3 \cdot 8 = 24 \). \( 24 = 5 + 19 \).

- If \( n = 16 \), then \( P(16) = 1 \cdot 6 = 6 \) and \( P(39) = 3 \cdot 9 = 27 \). \( 27 = 6 + 21 \).

- If \( n = 17 \), then \( P(17) = 1 \cdot 7 = 7 \) and \( P(40) = 4 \cdot 0 = 0 \). \( 0 \neq 7 + 23 \).

- If \( n = 18 \), then \( P(18) = 1 \cdot 8 = 8 \) and \( P(41) = 4 \cdot 1 = 4 \). \( 4 \neq 8 + 23 \).

- If \( n = 19 \), then \( P(19) = 1 \cdot 9 = 9 \) and \( P(42) = 4 \cdot 2 = 8 \). \( 8 \neq 9 + 23 \).

- If \( n = 20 \), then \( P(20) = 2 \cdot 0 = 0 \) and \( P(43) = 4 \cdot 3 = 12 \). \( 12 \neq 0 + 23 \).

- If \( n = 21 \), then \( P(21) = 2 \cdot 1 = 2 \) and \( P(44) = 4 \cdot 4 = 16 \). \( 16 \neq 2 + 23 \).

- If \( n = 22 \), then \( P(22) = 2 \cdot 2 = 4 \) and \( P(45) = 4 \cdot 5 = 20 \). \( 20 = 4 + 16 \).

- If \( n = 23 \), then \( P(23) = 2 \cdot 3 = 6 \) and \( P(46) = 4 \cdot 6 = 24 \). \( 24 = 6 + 18 \).

- If \( n = 24 \), then \( P(24) = 2 \cdot 4 = 8 \) and \( P(47) = 4 \cdot 7 = 28 \). \( 28 = 8 + 20 \).

- If \( n = 25 \), then \( P(25) = 2 \cdot 5 = 10 \) and \( P(48) = 4 \cdot 8 = 32 \). \( 32 = 10 + 22 \).

- If \( n = 26 \), then \( P(26) = 2 \cdot 6 = 12 \) and \( P(49) = 4 \cdot 9 = 36 \). \( 36 = 12 + 24 \).

- If \( n = 27 \), then \( P(27) = 2 \cdot 7 = 14 \) and \( P(50) = 5 \cdot 0 = 0 \). \( 0 \neq 14 + 23 \).

- If \( n = 28 \), then \( P(28) = 2 \cdot 8 = 16 \) and \( P(51) = 5 \cdot 1 = 5 \). \( 5 \neq 16 + 23 \).

- If \( n = 29 \), then \( P(29) = 2 \cdot 9 = 18 \) and \( P(52) = 5 \cdot 2 = 10 \). \( 10 \neq 18 + 23 \).

- If \( n = 30 \), then \( P(30) = 3 \cdot 0 = 0 \) and \( P(53) = 5 \cdot 3 = 15 \). \( 15 \neq 0 + 23 \).

- If \( n = 31 \), then \( P(31) = 3 \cdot 1 = 3 \) and \( P(54) = 5 \cdot 4 = 20 \). \( 20 = 3 + 17 \).

- If \( n = 32 \), then \( P(32) = 3 \cdot 2 = 6 \) and \( P(55) = 5 \cdot 5 = 25 \). \( 25 = 6 + 19 \).

- If \( n = 33 \), then \( P(33) = 3 \cdot 3 = 9 \) and \( P(56) = 5 \cdot 6 = 30 \). \( 30 = 9 + 21 \).

- If \( n = 34 \), then \( P(34) = 3 \cdot 4 = 12 \) and \( P(57) = 5 \cdot 7 = 35 \). \( 35 = 12 + 23 \).

Thus, the least positive integer \( n \) such that \( P(n+23) = P(n) + 23 \) is \(\boxed{34}\). 

\rule{\textwidth}{1pt} 

\textsc{\textbf{Sample 2}} \\ \\
\textbf{\textit{Question:}} \\ 
Consider the function $f(x)=x^{2}-2 x$. Determine all real numbers $x$ that satisfy the equation $f(f(f(x)))=3$. \\ \rule{\textwidth}{0.5pt}

\textbf{\textit{AceMath-72B-Instruct:}} \\
To solve the equation \( f(f(f(x))) = 3 \) where \( f(x) = x^2 - 2x \), we need to determine the values of \( x \) that satisfy this equation step by step.

First, let's find the values of \( x \) that satisfy \( f(x) = 3 \). 
We start with the equation:
\[
    x^2 - 2x = 3
\]
Rearranging terms, we get:
\[
    x^2 - 2x - 3 = 0
\]
This quadratic equation can be factored as:
\[
    (x - 3)(x + 1) = 0
\]
So, the solutions are:
\[
    x = 3 \quad \text{or} \quad x = -1
\]
This means that \( f(x) = 3 \) if and only if \( x = 3 \) or \( x = -1 \).

Next, we need to find the values of \( x \) such that \( f(f(x)) = 3 \). From the previous step, we know that \( f(f(x)) = 3 \) if and only if \( f(x) = 3 \) or \( f(x) = -1 \).

We already know the solutions to \( f(x) = 3 \), which are \( x = 3 \) and \( x = -1 \). Now, let's find the solutions to \( f(x) = -1 \):
\[
    x^2 - 2x = -1
\]
Rearranging terms, we get:
\[
    x^2 - 2x + 1 = 0
\]
This quadratic equation can be factored as:
\[
    (x - 1)^2 = 0
\]
So, the solution is:
\[
    x = 1
\]
Therefore, the solutions to \( f(f(x)) = 3 \) are \( x = 3 \), \( x = -1 \), and \( x = 1 \).

Finally, we need to find the values of \( x \) such that \( f(f(f(x))) = 3 \). From the previous step, we know that \( f(f(f(x))) = 3 \) if and only if \( f(f(x)) = 3 \) or \( f(f(x)) = -1 \).

We already know the solutions to \( f(f(x)) = 3 \), which are \( x = 3 \), \( x = -1 \), and \( x = 1 \). Now, let's find the solutions to \( f(f(x)) = -1 \):
\[
    f(f(x)) = -1
\]
From the previous step, we know that \( f(x) = 3 \) or \( f(x) = -1 \) or \( f(x) = 1 \) will satisfy \( f(f(x)) = 3 \). We already know the solutions to \( f(x) = 3 \), which are \( x = 3 \) and \( x = -1 \). We also know the solution to \( f(x) = -1 \), which is \( x = 1 \). Now, let's find the solutions to \( f(x) = 1 \):
\[
    x^2 - 2x = 1
\]
Rearranging terms, we get:
\[
    x^2 - 2x - 1 = 0
\]
This quadratic equation can be solved using the quadratic formula \( x = \frac{-b \pm \sqrt{b^2 - 4ac}}{2a} \), where \( a = 1 \), \( b = -2 \), and \( c = -1 \):
\[
    x = \frac{2 \pm \sqrt{4 + 4}}{2} = \frac{2 \pm \sqrt{8}}{2} = \frac{2 \pm 2\sqrt{2}}{2} = 1 \pm \sqrt{2}
\]
So, the solutions are:
\[
    x = 1 + \sqrt{2} \quad \text{or} \quad x = 1 - \sqrt{2}
\]
Therefore, the solutions to \( f(f(f(x))) = 3 \) are \( x = 3 \), \( x = -1 \), \( x = 1 \), \( x = 1 + \sqrt{2} \), and \( x = 1 - \sqrt{2} \).

The final answer is:
\[
    \boxed{3, -1, 1, 1 + \sqrt{2}, 1 - \sqrt{2}}
\]
\rule{\textwidth}{1pt}

\textsc{\textbf{Sample 3}} \\ \\
\textit{\textbf{Question:}} \\
Alice chooses a set $A$ of positive integers. Then Bob lists all finite nonempty sets $B$ of positive integers with the property that the maximum element of $B$ belongs to $A$. Bob's list has 2024 sets. Find the sum of the elements of A. \\
\rule{\textwidth}{0.5pt}

\textit{\textbf{AceMath-72B-Instruct:}} \\
To solve the problem, we need to determine the set \( A \) such that the number of finite nonempty sets \( B \) of positive integers, where the maximum element of \( B \) belongs to \( A \), is 2024. Let's denote the elements of \( A \) as \( a_1, a_2, \ldots, a_n \) where \( a_1 < a_2 < \cdots < a_n \).

For each element \( a_i \) in \( A \), the sets \( B \) that have \( a_i \) as their maximum element are all subsets of \( \{1, 2, \ldots, a_i\} \) that include \( a_i \). The number of such subsets is \( 2^{a_i-1} \) because we can either include or exclude each of the \( a_i-1 \) elements \( 1, 2, \ldots, a_i-1 \).

Therefore, the total number of sets \( B \) is the sum of \( 2^{a_i-1} \) for all \( i \) from 1 to \( n \):
\[
2^{a_1-1} + 2^{a_2-1} + \cdots + 2^{a_n-1} = 2024.
\]

We need to find a set \( A \) such that the above sum equals 2024. Notice that the sum is a sum of distinct powers of 2, which means it is the binary representation of 2024. Let's convert 2024 to binary:
\[
2024_{10} = 11111101000_2.
\]
This binary number can be expanded as:
\[
2024 = 2^{10} + 2^9 + 2^8 + 2^7 + 2^6 + 2^5 + 2^3.
\]
Thus, the corresponding set \( A \) is:
\[
A = \{11, 10, 9, 8, 7, 6, 4\}.
\]

The sum of the elements of \( A \) is:
\[
11 + 10 + 9 + 8 + 7 + 6 + 4 = 55.
\]

Therefore, the sum of the elements of \( A \) is \(\boxed{55}\).

\rule{\textwidth}{1pt}

\textsc{\textbf{Sample 4}} \\ \\
\textit{\textbf{Question:}} \\
How many 'r' in 'strawberry'? \\
\rule{\textwidth}{0.5pt}

\textit{\textbf{AceMath-72B-Instruct:}} \\
To determine how many 'r' letters are in the word "strawberry," we can follow these steps:

1. Write down the word: "strawberry."

2. Identify each 'r' in the word.

3. Count the number of 'r's.

Let's go through the word "strawberry" step by step:

- The first letter is 's'.

- The second letter is 't'.

- The third letter is 'r'.

- The fourth letter is 'a'.

- The fifth letter is 'w'.

- The sixth letter is 'b'.

- The seventh letter is 'e'.

- The eighth letter is 'r'.

- The ninth letter is 'r'.

- The tenth letter is 'y'.

Now, let's count the 'r's:

- There is one 'r' in the third position.

- There is one 'r' in the eighth position.

- There is one 'r' in the ninth position.

So, the total number of 'r's in the word "strawberry" is \(1 + 1 + 1 = 3\).

Therefore, the number of 'r' in 'strawberry' is \(\boxed{3}\).

\rule{\textwidth}{1pt}

\end{document}